\documentclass[10pt,journal]{IEEEtran}
\usepackage{graphicx}
\usepackage{times}
\usepackage{epsfig}
\usepackage{amsmath}
\usepackage{amssymb}
\usepackage{overpic}
\usepackage{subfigure}
\usepackage{multirow}
\usepackage{color}
\usepackage{cite}
\usepackage{diagbox}
\usepackage{bigstrut}
\usepackage{booktabs}
\usepackage{animate}

\usepackage[numbers,sort&compress]{natbib}

\usepackage{color,soul}

\renewcommand{\raggedright}{\leftskip=0pt \rightskip=0pt plus 0cm}

\definecolor{darkcyan}{rgb}{0.0, 0.55, 0.55}
\newcommand{\wh}[1]{{\color{black} #1}}
\newcommand{\wht}[1]{{\color{black} #1}}
\newcommand{\whtt}[1]{{\color{black} #1}}
\newcommand{\ly}[1]{{\color{black} #1}}
\usepackage[colorlinks,linkcolor=blue]{hyperref}
\def \useAnimate {}

\begin{document}
\title{Video Coding for Machines: A Paradigm of Collaborative Compression and Intelligent Analytics}

\author{Ling-Yu Duan, Jiaying Liu, Wenhan Yang, Tiejun Huang, Wen Gao
\thanks{L.-Y. Duan, J. Liu, W. Yang, T. Huang, and W. Gao are with Peking University, Beijing 100871, China (e-mail: 	lingyu@pku.edu.cn; liujiaying@pku.edu.cn; yangwenhan@pku.edu.cn; tjhuang@pku.edu.cn; wgao@pku.edu.cn).}}

\markboth{}
{Shell \MakeLowercase{\textit{et al.}}: Bare Demo of IEEEtran.cls for Computer Society Journals}

\maketitle
	
\begin{abstract}
Video coding, which targets to compress and reconstruct the whole \whtt{frame}, and \whtt{feature} compression, which only preserves and transmits the most critical information, stand at two ends of the scale.
\whtt{That is}, one \ly{is} with compactness and efficiency to serve for machine vision, and the other \ly{is} with full fidelity, bowing to human \whtt{perception}.
The recent endeavors in \whtt{imminent trends of video compression}, \textit{e.g.} \whtt{deep learning} based coding tools and end-to-end image/video coding, and MPEG-7 compact feature descriptor standards, \textit{i.e.} Compact Descriptors for Visual Search and Compact Descriptors for Video Analysis, promote the sustainable and fast development in their own directions, respectively. 
In this paper, thanks to \whtt{booming} AI \whtt{technology}, \textit{e.g.} \whtt{prediction} and generation models, we \whtt{carry out exploration in the new area, Video Coding for Machines}~(VCM), arising from the emerging MPEG standardization efforts\footnote{https://lists.aau.at/mailman/listinfo/mpeg-vcm}. Towards collaborative compression and intelligent analytics, VCM attempts to bridge the gap between feature coding for machine vision and video coding for human vision.
Aligning with \whtt{the rising} \whtt{\textit{Analyze then Compress}} instance Digital Retina, the definition, formulation, and paradigm of VCM are given first.
\whtt{Meanwhile,} we systematically review state-of-the-art techniques in video compression and feature compression from the unique perspective of MPEG standardization, which provides the academic and industrial evidence \whtt{to realize} the collaborative compression of video and feature streams in a broad range of AI applications. Finally, we come up with potential VCM solutions, and the preliminary results have demonstrated the performance and efficiency gains. Further direction is discussed as well. 
\end{abstract}



\IEEEpeerreviewmaketitle

\renewcommand{\baselinestretch}{0.95}

\section{Introduction}
\label{sec:introduction}

In the big data era, \wh{massive videos are fed into machines to realize intelligent analysis in numerous applications of smart cities or Internet of things (IoT). 
\whtt{Like} the explosion of surveillance systems deployed in urban areas, there arise important concerns on how to efficiently manage \whtt{massive} video data. There is a unique set of challenges (\textit{e.g.} low latency and high accuracy) regarding \ly{efficiently} \whtt{analyzing and searching the target within the}
millions of objects/events captured everyday. In particular, video compression and transmission constitute the basic infrastructure to support these applications from the perspective of \whtt{\textit{Compress then Analyze}}.} Over the past decades, a series of \wh{standards} \wh{(\whtt{\textit{e.g.}} MPEG-4 AVC/H.264~\cite{H264} and High Efficiency Video Coding (HEVC)~\cite{H265})}, Audio Video coding Standard (AVS)~\cite{ma2015avs2} are built to significantly improve the \wh{video} coding efficiency, by
\wh{squeezing out the spatial-temporal pixel-level redundancy of video frames based on the visual signal statistics and the priors of human perception.} \wh{More recently, deep learning based video coding makes great progress.}
With the hierarchical model architecture and the \ly{large-scale} data priors, these methods \ly{largely outperform} the state-of-the-art codecs \wh{by} utilizing deep-network aided coding tools. 
\wh{Rather than directly targeting machines,} \wh{these} methods 
\wh{focus on efficiently reconstructing} the pixels for human vision, in which the spatial-temporal volume of pixel arrays can be fed into machine learning \wh{and} pattern recognition algorithms to complete high-level analysis \wh{and retrieval} tasks.

However, when facing big data and video analytics, existing video coding methods (even for the \whtt{deep learning based}) \wh{are still questionable, regarding whether such big video data can be efficiently handled by visual signal level compression}.
\wh{Moreover,} the full-resolution videos \wh{are} of low density in practical values.
It is \wh{prohibitive} to compress and store all video data first and then perform analytics \wh{over the decompressed} video stream. \wh{By degrading the quality of compressed videos, \whtt{it might} save more bitrates, but incur the risk of degraded analytics performance due to the poorly extracted features.}

To facilitate the high-level machine vision tasks in terms of performance and efficiency, lots of research efforts have been dedicated to extracting those pieces of key information, \textit{i.e.}, visual features, from the pixels, which \whtt{is} usually compressed and represented in a very compact form. 
\wh{This poses an alternative strategy \textit{Analyze then Compress}, which extracts, saves, and transmits compact features to satisfy various intelligent video \whtt{analytics} tasks, by using significantly less data than the compressed video itself. In particular, to meet the demand for large-scale video analysis in smart city applications, the feature stream instead of the video signal stream can be transmitted. In view of the necessity and importance of transmitting feature descriptors, MPEG has finalized the standardization of compact descriptors for visual search (CDVS) (ISO/IEC15938-13) in Sep. 2015~\cite{cdvs} and compact descriptors for video analysis (CDVA) (ISO/IEC15938-15)~\cite{cdva} in \whtt{July} 2019 to enable the interoperability for efficient and effective image/video retrieval and analysis by standardizing the bitstream syntax of compact feature descriptors. In CDVS, hand-crafted local and global descriptors are designed to represent the visual characteristics of images. In CDVA, the deep learning features are adopted to further boost the video analysis performance.}
\wh{Over the course of the standardization process, remarkable improvements \whtt{are} achieved in reducing the size of features while maintaining their discriminative power for machine vision tasks. Such compact features cannot reconstruct the full resolution videos for human observers, thereby incurring two successive stages of analysis and compression for machine and human vision.}

\wh{For either \textit{Compress then Analyze} or \textit{Analyze then Compress}, the optimization jobs of video coding and feature coding are separate.
Due to the very nature of multi-tasks for machine vision and human vision, the intrusive setup of two separate stages is sub-optimal. It is expected to explore more collaborative operations between video and feature streams, which opens up more space for improving the performance of intelligent \whtt{analytics}, optimizing the video coding efficiency, and thereby reducing the total cost.
Good opportunities to bridge the cross-domain research on machine vision and human vision have been there, as deep neural network has demonstrated its excellent capability of multi-task end-to-end optimization as well as abstracting the \whtt{representations} of multiple granularities in a hierarchical architecture.
}

\wh{In this paper, we attempt to identify the opportunities and challenges of developing collaborative compression techniques for humans and machines. Through reviewing the \ly{advance} of two separate tracks of video coding and feature coding, we present the necessity of machine-human collaborative compression, and formulate a new problem of video coding for machines (VCM). Furthermore, to promote the emerging MPEG-VCM standardization efforts and collect for evidences for MPEG \whtt{Ad hoc} Group of VCM, we \ly{propose} \whtt{\ly{trial}} exemplar VCM architectures and conduct preliminary experiments, which are also expected to provide \ly{insights} into bridging the cross-domain research from visual signal processing, computer vision, and machine learning, when AI meets the video big data. The contributions are summarized as follows,} 

\wh{
\begin{itemize}
	\item 
	We present and formulate a new problem of Video Coding for Machines by identifying three elements of \whtt{\textit{tasks, features}, and \textit{resources}}, in which \ly{a} novel feedback mechanism for \whtt{\textit{collaborative} and \textit{scalable}} modes is introduced to improve the coding efficiency and the analytics performance for both human \whtt{and} machine-oriented tasks.
	\item 
	We review the state-of-the-art approaches in video compression and feature compression from a unique perspective of standardized technologies, and study the impact of more recent deep image/video prediction and generation models on the potential VCM related techniques.
	\item 
	We propose exemplar VCM architectures and provide potential  solutions. \ly{Preliminary} experimental results have shown the advantages of VCM collaborative compression in improving video and feature coding efficiency and performance for human and machine vision. 
\end{itemize}
}
\vspace{1mm}

\whtt{The rest of the article is organized as follows.
Section~\ref{sec:video_compression} and~\ref{sec:feat_compression} briefly review previous works on video coding and feature compression, respectively.
Section~\ref{sec:vcm_define} provides the definition, formulation, \whtt{and paradigm} of VCM.
Section~\ref{sec:trend} illustrates \whtt{the} emerging \whtt{AI technique}, \whtt{which provides useful evidence} 
for VCM,
After that, in Section~\ref{sec:results}, we \ly{provide} potential solutions for \wh{VCM} problem.
In Section~\ref{sec:results}, the preliminary experimental results are reported.
In Section~\ref{sec:discussions}, several issues, \whtt{and} future directions are discussed.
In Section~\ref{sec:conclusion}, the concluding remarks are provided.}

\section{Review of Video Compression: From Pixel Feature Perspective}
\label{sec:video_compression}
Visual information takes up at least 83\% of all information~\cite{visual_perception} that people can feel.
\whtt{It} is important for humans to record, store, and view the image/videos efficiently. \wh{For past decades, lots of academic and industrial efforts have been devoted to video \whtt{compression}, which is to maximize the \whtt{compression} efficiency from the pixel feature perspective. Below we review the advance of \whtt{traditional} video coding as well as the impact of deep learning based \whtt{compression} on visual data coding for human vision \whtt{in} a general sense.}
\vspace{-2mm}

\subsection{Traditional Hybrid Video Coding}

Video coding transforms the input video into a compact binary code for more economic and light-weighted storage and transmission, and targets \wh{reconstructing} videos visually \wh{by} the decoding process.
In 1975, the \textit{hybrid spatial-temporal coding architecture}~\cite{Hybrid_coding} is proposed to take the lead and occupy the major proportion during the next few decades.
After that, the following video coding standards have evolved through the \whtt{development} of the ITU-T and ISO/IEC standards.
The ITU-T produced H.261~\cite{261} and 
H.263~\cite{263}, 
ISO/IEC produced MPEG-1~\cite{MPEG1}
and MPEG-4 Visual~\cite{MPEG4},
and the two organizations worked together to produce the H.262/MPEG-2 Video~\cite{Generic_Coding}, 
H.264/MPEG-4 Advanced Video Coding (AVC)~\cite{Advanced_Coding} standards,
and H.265/MPEG-H (Part 2) High Efficiency Video Coding~(HEVC)~\cite{HEVC} standards.

The design \ly{and development} of all these standards follows the \textit{block-based \whtt{video} coding} approach.
The first \wh{technical} feature is block-wise partition.
Each coded picture is partitioned into macroblocks (MB) of luma and chroma samples.
\whtt{MBs} will be divided into slices \whtt{and} coded independently.
Each slice is further partitioned into coding tree units.
After that, the coding unit (CU), prediction unit (PU), and transform unit (TU) are obtained to make the coding, prediction and transform \wh{processes} more \ly{flexible}.
Based on the block-based design, the intra and inter-predictions are applied based on PU and the corresponding contexts, \textit{i.e.} neighboring blocks and reference frames in the intra and inter modes, respectively.
\whtt{But these kinds of designed patterns}
\ly{just cover} parts of the context information, which limits the modeling capacity in prediction.
\ly{Moreover}, the block-wise prediction, along with transform and quantization, leads to the discontinuity at the block boundaries.
With the quantization of the residue or original signal in the transform block, the blockness \ly{appears}.
To suppress the artifacts, the loop filter is \ly{applied for smoothing.}

Another important \wh{technical} \wh{feature is} \whtt{\textit{hybrid video coding}}.
Intra and inter-prediction are used to remove temporal and spatial statistical redundancies, respectively.
For intra-prediction, HEVC \whtt{utilizes} a line of preceding reconstructed pixels above and on the left side of the current PU as the reference for generating predictions. 
The number of intra modes is 35, including planar mode, DC mode, and 33 angular modes.
It is performed in the transform unit.
For inter-prediction, HEVC derives a motion-compensated prediction  for a block of
image samples.
\ly{The} homogeneous motion inside a block \ly{is assumed, and the size of a moving object is usually larger than one block.}
Reference blocks will be searched from previously coded pictures for inter prediction.
For both intra and inter-predictions, 
\whtt{the best mode \ly{is selected by} the} Rate-Distortion Optimization \whtt{(RDO)~\cite{RDO}}.
However, the multi-line prediction scheme and the block-wise reference block might not provide a desirable prediction reference when the structures and motion are complex.
Besides, 
\ly{when the RDO boosts the modeling capacity, the overhead of signaled bits and computation occur.}
\ly{The target of optimizing the rate distortion efficiency is} to \ly{seek for} the trade-off in bitrate and signal distortion. It can be \ly{solved} via Lagrangian optimization techniques. Coder control \wh{finally} determines a set of coding parameters that affect the encoded bit-streams.

With tremendous expert efforts, the coding performance is \ly{dramatically} improved in the past decade that 
there is the rule of thumb that, one generation of video coding standards almost surpasses the previous one by up to 50\% in coding efficiency.

\subsection{Deep Learning Based Video Coding}
\wh{The} \whtt{deep learning} techniques significantly promote the development of video \whtt{coding.}
The seminar work~\cite{2015arXiv151106281B} in 2015 opened a door to the end-to-end learned video coding. 
\wh{The} \whtt{deep learning} based coding tools~\cite{first_loop_filter,deep_partition} are developed since 2016. Benefiting from the bonus of big data, \ly{powerful} architectures, end-to-end optimization, and \ly{other} advanced techniques, \textit{e.g.} unsupervised learning,
\wh{the emerging} deep learning \ly{excel in learning} data-driven priors for \ly{effective} video coding.
First, \ly{complex} nonlinear mappings can be modeled by \ly{the} hierarchical structures of neural networks, which \ly{improves} prediction \ly{models} to make the reconstructed signal more similar to the original one.
Second, deep structures, such as PixelCNN and PixelRNN, are capable to model the pixel probability and provide \ly{powerful generation functions} to model the visual signal in a more compact form.

\ly{The deep learning based coding methods} do not rely on the partition scheme and support full resolution coding, \ly{and thereby removing the blocking artifacts naturally.}
\ly{Since the partition is not required,} these modules can \wh{access} more context \ly{in} \ly{a larger region}.
\ly{As the features are} extracted via a hierarchical network and jointly optimized with the \wh{reconstruction task}, \ly{the resulting features tend to \ly{comprehensive} and powerful for high efficient coding.}
More efforts are put into increasing the receptive field \ly{for better perception} of larger regions via recurrent neural network~\cite{8100060,2015arXiv151106085T} and nonlocal attention~\cite{Liu_2019_CVPR_Workshops,2019arXiv190409757L}, leading to improved coding performance.
The former infers the latent representations of image/videos progressively. In each iteration, with the previously \ly{perceived} context, the network removes the unnecessary bits from the latent representations to \ly{achieve more compactness}.
The \ly{latter} makes efforts to \ly{figure out} the nonlocal correspondence among endpixels/regions, \ly{in order to} remove the long-term spatial redundancy.

The efforts \ly{in improving} the performance of neural network-aided coding tools rely on the excellent prediction ability of deep networks.
\ly{Many works attempt to} effectively learn the \wh{end-to-end} mapping \ly{in} a series of video coding tasks, \textit{e.g.},
intra-prediction~\cite{multiple_lines,progressive_RNN},
\whtt{inter}-prediction~\cite{inter_xia,inter_zhao,Frac_Interp,one_for_all,invertible_interp},
deblocking~\cite{DMCNN,chuanmin_loop_filter,PRN,loop_filter_cu_class}, and fast mode decision~\cite{cnn_cu_mode}.
With a \ly{powerful} and unified prediction model, these methods obtain superior R-D performance.
\ly{For} intra-prediction, the diversified modes \ly{derived from} RDO in the traditional video codec \wh{are} replaced by a learned general model given a certain kind of context.
In~\cite{multiple_lines}, based on the multi-line reference context,
fully-connected (FC) neural networks are utilized to \ly{generate} the prediction result.
In~\cite{Context_prediction}, the \ly{strengths} of FC and CNN networks are combined.
In~\cite{progressive_RNN}, benefiting from the end-to-end learning and the block-level reference scheme, the \ly{proposed} predictor \ly{employs context information in a large scale to suppress quantization noises}.
\ly{For} inter-prediction, deep network-based methods~\cite{deep_inter,8451465} break the limit of block-wise motion compensation and bring in better inter-frame prediction, namely \ly{generating better reference frames by using all the reconstructed frames.}
\ly{For} loop-filter, the \ly{powerful} network architectures and the full-resolution input of all reconstructed frames~\cite{vcip18_sota,drn} \ly{significantly improve the performance of loop filters, say} up to 10\% BD-rate reduction \ly{as reported} in many methods.

The end-to-end \ly{learning based} compression \ly{leverages deep networks} to model the pixel distribution and generate complex signals from a very compact representation.
The pioneering work~\cite{2015arXiv151106085T} proposes a parametric nonlinear transformation \wh{to} well {Gaussianize} data \ly{for reducing} mutual information between transformed components and show
impressive \ly{results in} image compression.
Meanwhile, in~\cite{2015arXiv151106281B}, a general framework is built upon convolutional and deconvolutional LSTM recurrent networks, trained once,
to support variable-rate image compression via reconstructing the image progressively.
Later works~\cite{8100060,2016arXiv160705006B,2016arXiv161101704B,2018arXiv180201436B} \ly{continue to improve compression efficiency by following the routes}.
All these methods \ly{attempt to reduce} the overall R-D cost on a large-scale dataset.
Due to the model's flexibility, there are also more \ly{practical} ways to control the bitrate adaptively, such as applying the attention mask~\cite{8578437} to guide the use \wh{of} more bits on complex regions.
Besides, \wh{as} the R-D cost is optimized in an end-to-end manner, it is \ly{flexible} to 
\wh{adapt the rate and distortion to accommodate a variety of end applications, \textit{e.g.} machine vision tasks.}

Although video coding performance is improved \wh{constantly},
\wh{some intrinsic problems exist}\ly{,} especially when tremendous volumes of data need to be processed and analyzed\ly{.}
The low-value data volume \wh{still constitutes a major} \ly{part.}
\wh{So these methods \ly{of reconstructing whole pictures} cannot fulfill the requirement of real-time video content analytics when dealing with large-scale video data.}
However, the \ly{strengths} in \whtt{deep learning} based image/video coding, \textit{i.e.} \ly{the} excellent prediction and generation capacity of deep models and the flexibility of R-D cost optimization, provide opportunities to develop \ly{VCM technology} \wh{to address these challenges}.

\section{Review of Feature Compression: From Semantic Feature Perspective}
\label{sec:feat_compression}

\wh{The traditional video coding targets high visual fidelity for humans. With the proliferation of applications that capture video for (remote) consumption by a machine, such as connected vehicles, video surveillance systems, and video capture for smart cities, more recent efforts on feature compression target low bitrate intelligent \whtt{analytics} (\textit{e.g.}, image/video recognition, classification, retrieval) for machines, as transmission bandwidth for raw visual features is often at a premium, even for the emerging strategy of \textit{Analyze then Compress}.}

\wh{However, it is not new to explore feature descriptors for MPEG. Back in 1998, MPEG initiated MPEG-7, formally known as Multimedia Content Description Interface, driven by the needs for tools and systems to index, search, filter, and manage audio-visual content. Towards interoperable interface, MPEG-7~\cite{visual_shape} defines the syntax and semantics of various tools for describing color, texture, shape, motion, \textit{etc}. Such descriptions of streamed or stored media help human or machine users to identify, retrieve, or filter audio-visual information. Early visual descriptors developed in MPEG-7 have limited usage, as those low-level descriptors are sensitive to scale, rotation, lighting, occlusion, noise, \textit{etc}. More recently, the advance of computer vision and deep learning has significantly pushed forward the standardized visual descriptions in MPEG-7.  In particular, CDVS~\cite{MPEG_CDVS} and CDVA~\cite{MPEG_CDVA} have been in the vanguard of the trend of \textit{Analyze then Compress} by extracting and transmitting compact visual descriptors.}

\begin{figure*}[t]
	\centering
	\includegraphics[width=0.9\linewidth]{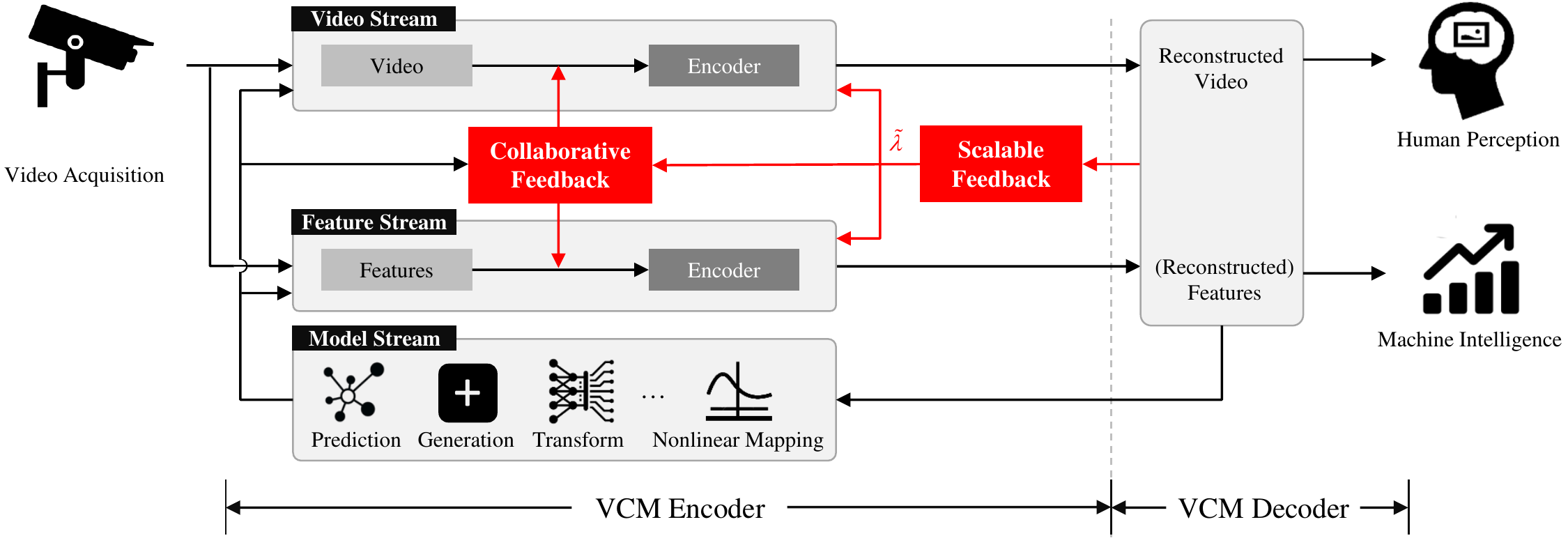}
	\caption{The proposed video coding for machines (VCM) framework by incorporating the feedback mechanism into the collaborative coding of multiple streams including videos, features, and models, targeting the multi-tasks of human perception and machine intelligence.}
	\label{fig:VCM_framework}
\end{figure*}

\subsection{CDVS}

\wh{CDVS can trace back to the requirements for early mobile visual search systems by 2010, such as faster search, higher accuracy, and better user experience. Initial research~\cite{mobile,CHG,mobile_efficient,tree_histogram_coding,low_bit1,low_bit2} demonstrated that one could reduce transmission data by at least an order of magnitude by extracting compact visual features on the mobile device and sending descriptors at low bitrates to a remote machine for search.
\ly{Moreover, a} significant reduction in latency could also be achieved when performing all processing on the device itself. Following initial research, an exploratory activity in the MPEG was initiated at the 91\textit{st} meeting (Kyoto, Jan. 2010). In July 2011, MPEG launched the standardization of CDVS. 
\ly{The} CDVS standard (formally known as MPEG-7, Part 13) was published by ISO on \whtt{Aug. 25}, 2015, which specifies a normative bitstream of standardized \ly{compact visual} descriptors for mobile visual search and augmented reality applications. 

Over the course of the standardization process, CDVS has made remarkable improvements \ly{over a large-scale benchmark} in image matching/retrieval performance with very compact feature descriptors (at \whtt{six} predefined descriptor lengths: 512B, 1KB, 2KB, 4KB, 8KB, and 16KB)~\cite{cdvs}. High performance is achieved while stringent memory and computational complexity requirements are satisfied to make the standard ideally suited for both hardware and software implementations~\cite{cdvs_gpu}.

To guarantee the interoperability, CDVS makes a normative feature extraction pipeline including interest point detection, local feature selection, local feature description, local feature descriptor aggregation, local feature descriptor compression, and local feature location compression.
Key techniques, \textit{e.g.} low-degree polynomial detector~\cite{Response_CDVS,accurate_efficient,MPEG_CDVS}, light-weighted interest point selection~\cite{Interest_Point,FRANCINI2013311}, scalable compressed fisher vector~\cite{codebook_free,rate_adaptive_compact_fisher}, and location histogram coding~\cite{location_coding,Tsai2012Improved}, have been developed by competitive and collaborative experiments within a rigorous evaluation framework~\cite{framework_CDVS}.

It is worthy to mention that CDVS makes a normative encoder (the feature extraction process is fixed), which is completely distinct from conventional video coding standards in making a normative decoder. The success of CDVS standardization originates from the mature computer vision algorithm (\whtt{like} SIFT~\cite{sift}) on the reliable image matching between different views of an object or scene. However, when dealing with more complex analysis tasks in the autonomous drive and video surveillances, the normative encoder process incurs more risks of lower performance than those task-specific fine-tuned features directly derived from an end-to-end learning framework. Fortunately, the collaborative compression of video and feature streams is expected to address this issue, as we may leverage the joint advantages of the normative feature encoding process and the normative video decoding process.}

\subsection{CDVA}

\wh{The bold idea of CDVA initiated in the 111\textit{th} MPEG meeting in \whtt{Feb.} 2015 is to have a normative video feature descriptor based on neural networks for \ly{machine vision tasks}, targeting an exponential increase in the demand for video analysis in autonomous drive, video surveillance systems, entertainment, and smart cities. The standardization at the encoder requires the deterministic deep network model and parameters. However, due to fast-moving deep learning techniques and their end-to-end optimization nature, there is a lack of a generic deep model \ly{sufficing for} a broad of video analytics tasks. This is the primary challenge for CDVA.

To kick off this standardization, CDVA narrows down to the general task of video matching and retrieval, aims at determining if a pair of videos share the object or scene with similar content, and searching for videos containing similar segments to the one in the query video. Extensive experiments~\cite{HNIP} over CDVA benchmark report comparable performances of the deep learning features with different off-the-shelf CNN models (\whtt{like} VGG16~\cite{vgg_19}, AlexNet~\cite{Krizhevsky_2012}, ResNet~\cite{resNet}), which \whtt{provide} useful evidence for normative deep learning features for the task of video matching and retrieval. Due to the hardware friendly merits of uniformly sized small filters (3x3 convolution kernel and 2x2 max pooling), and the competitive performance of combining convolution layers of the small filters by replacing a large filter (5 $\times$ 5 or 7 $\times$ 7), VGG16 is adopted by CDVA as the normative backbone network to derive compact deep invariant feature representations.

The CDVA standard (formally known as MPEG-7, Part 15) was published by ISO in July 2019. Based on a previously agreed-upon methodology, key technologies are developed by MPEG experts, including Nested Invariance Pooling (NIP)~\cite{HNIP} for deep invariance global descriptor, Adaptive Binary Arithmetic Coding (ABAC) based temporal coding of global and local descriptors~\cite{MPEG_CDVA}, the integration of hand-crafted and deep learning features~\cite{cdva}. 
\ly{The NIP method produces compact global descriptors from a CNN
model by progressive pooling operations to improve the translation, scale and rotation invariance over the feature maps of intermediate network layers.}
The extensive experiments have demonstrated that the NIP (derived from the last pooling layer, \textit{i.e.}, pool5, of VGG16) outperforms the state-of-the-art deep and canonical hand-crafted descriptors with significant gains. In particular, the combination of (deep learning) NIP global descriptors and (hand-crafted) CDVS global descriptors has significantly boosted the performance with a comparable descriptor length.

More recently, great efforts are made to extend the CDVA standard for \ly{VCM}. Smart tiled CDVA~\cite{Smart_CDVA} \ly{targeting} video/image analysis of higher resolution input, can be applied to rich tasks including fine-grained feature extraction network, small object detection, segmentation, and other use cases. 
Smart sensing~\cite{Smart_Sensing} is an extension of the CDVA standard to directly process raw Bayer pattern data, without the need of a traditional ISP. This enables all Bayer pattern sensors to be natively CDVA compatible. SuperCDVA~\cite{Super_CDVA} is an application of the CDVA standard for video understanding by using the temporal information, in which each CDVA vector is embedded as one block in an image, and the multiple blocks are put in a sequential manner to represent the sequence of video frames for classification by a 2-D CNN model.

In summary, the new MPEG-7 standard CDVA opens the door to exciting new opportunities in supporting machine-only encoding as well as hybrid (machine and human) encoding formats to unburden the network and storage resources for dramatic benefits in various service models and a broad range of applications with \ly{low latency and reduced cost}.}

\section{Video Coding for Machine\wh{s}: Definition, Formulation and Paradigm}
\label{sec:vcm_define}
\wh{The emerging requirements of video processing and content analysis in a collaborative manner are expected to support both human vision and machine vision.}
\wh{In practice, two separate streams of compressed video data and compact visual feature descriptors are involved to satisfy the human observers and a variety of low-level and high-level \ly{machine} vision tasks, respectively.}
\wh{How to further improve the visual signal compression efficiency as well as the visual task performance by leveraging multiple granularities of visual features remains a challenging but promising task, in which the image/video data is considered as a sort of pixel-level feature.}

\subsection{Definition and Formulation}
\wh{\whtt{Traditional} \textit{video coding} targets visual signal fidelity at high \wh{bitrates}, while \textit{feature coding} targets high performance of vision tasks at very low \wh{bitrates}.}
\wh{Like CDVS and CDVA, most of the existing feature coding approaches heavily rely on specific tasks, which significantly save bitrate by ignoring the requirement of full-resolution video reconstruction.}
\wh{By contrast, the video codecs like HEVC \whtt{focus} on the reconstruction of full-resolution pictures from the compressed bit stream.
Unfortunately, the pixel-level features from data decompression cannot suffice for large-scale video analysis in a huge or a moderate scale camera network efficiently, due to the bottleneck of computation, communication, and storage~\cite{smart_visual_sensing}.}

\wh{We attempt to identify the role of \ly{Video Coding for Machines}~(\textbf{VCM}), as shown in Fig.~\ref{fig:VCM_framework}, to bridge the gap between coding semantic features for machine vision tasks and coding pixel features for human vision.}
\whtt{
The scalability is meant to incrementally improve the performance in a variety of vision tasks as well as the coding efficiency by optimizing bit utility between multiple low-level and high-level feature streams.}
\wh{Moreover, VCM is committed to developing key techniques for economizing the use of a bit budget to collaboratively complete multiple tasks targeting humans and/or machines.}

\wh{VCM aims to \textit{jointly maximize the performance of multiple tasks ranging from low-level processing to high-level semantic analysis, but minimize the use of  communication and computational resources}.}
\wh{VCM relates to three key elements:}
\begin{itemize}
	\item \textit{\wh{Tasks}}.
	VCM \wh{incurs low-level signal fidelity as well as high-level semantics, which may request a complex optimization objective from multiple tasks. It is important to figure out an efficient coding scheme across tasks.}
	\item \textit{Resources}. \wh{VCM is committed to tackling the practical performance issue, which cannot be well solved by traditional video coding or feature coding approaches solely, subject to the constraints of resources like bandwidth, computation, storage, and energy.}
	\item \textit{Features}. \wh{VCM is to explore a suite of rate distortion functions and optimization solutions from a unified perspective of leveraging the features at different granularities including the pixels for humans to derive efficient compression functions (\textit{e.g.}, transform and prediction).}
\end{itemize}

\wh{Here we formulate the VCM problem.}
The features are denoted by
$\mathbf{F} = \left \{ F^{0}, F^{1}, ..., F^{z} \right\} $ of \whtt{$(z+1)$} \wh{tasks}, where \whtt{$V:=F^{z}$} is the pixel feature, 
\wh{associated with the task of improving visual signal fidelity for human observers.} $F^{i}$ denotes \wh{more task (non-)specific semantic or syntactic features if $0 \le i  < z$.
A smaller $i$ denotes that the feature is more abstract.}
The performance of the \wh{task} $i$ is defined as follows:
\begin{equation}
\label{eq:task_performance}
q^{i} =  \Phi^{i}(\hat{F}^{i}),
\end{equation}
where $\Phi^{i}(\cdot)$ is the \wh{quality} metric \wh{related to task} $i$ and $\hat{F}^{i}$ \wh{is the reconstructed feature undergoing the encoding and decoding processes.
Note that, certain compressed domain processing or analysis may not require the complete reconstruction process.} 
We use $C(\cdot | \theta_c ), D(\cdot | \theta_d)$, and $\mathcal{G}(\cdot| \theta_g)$ to denote the processes of compression, decompression, feature \wh{prediction}, where $\theta_c$, $\theta_d$, and $\theta_g$ are parameters of the corresponding processes.
We define $S(\cdot)$ to \wh{measure the resource usage of computing a given feature, namely compressing, transmitting, and storing the feature, as well as further analyzing the feature.}
\wh{By incorporating the elements of tasks, resources, and features, VCM explicitly or non-explicitly  models a complex objective function for maximizing the performance and minimizing the resource cost, in which joint optimization applies.}
\wh{Generally}, the \textbf{VCM optimization function} is described as follows,
\begin{align}
\label{eq:vcm1}
& \mathop {\text{argmax} }\limits_{\Theta  = \left\{ {{\theta _c},{\theta _d},{\theta _g}} \right\}} \sum\limits_{0 \le i \le z} {{\omega ^i}{q^i}}, \nonumber \\
&\text{ subject to:} \sum\limits_{0 \le i \le z} {{\omega ^i}}  = 1, \\
 S\left( {{R_{{F^0}}}} \right) + & \sum  \limits_{i > 0}^{} {\mathop {\min }\limits_{0 \le j < i} } \left\{ {S\left( {{R_{{F_{i \to j}}}}} \right)} \right\} + S\left( {{R_M}} \right) + S\left( \Theta  \right) \le {S_T}, \nonumber 
\end{align}
where ${S_T}$ is total resource cost constraint, and ${\omega ^i}$ balances the importance of different tasks in the optimization.
The first two terms in the resource constraint are the resource cost of compressing and transmitting features of all tasks with feature prediction. Note that,
${R_V} = {\min _j}\left\{ {S\left( {{R_{{F_{z \to j}}}}} \right)} \right\}$ is the resource cost to encode videos \whtt{${F^z}$}  with feature prediction.
The third term in the resource constraint is \wh{the resource overhead from leveraging models and updating models accordingly.}
\wh{The last term $S\left( \Theta  \right)$ calculates the resource cost caused by using a more complex model, 
such as longer training time,
response time delay due to feedbacks,
larger-scale training dataset.		
Principle rules apply to the key elements as below,}
\begin{align}
{R_{{F^0}}} & = C\left( {{F^0}|{\theta _c}} \right),  \\
{R_{{F_{i \to j}}}} & = C\left( {F^i} - {\rm{{\cal G}}}\left( {{F^j},i|{\theta _g}} \right)\right ), \text{ }{\rm{ for }} \text{ } i \ne 0,  \label{eq:pred} \\
{\widehat F^0} & = D\left( {{R_{{F^0}}}|{\theta _d}} \right), \\
{\widehat F^i} & = D\left( {{R_{{F_{i \to j}}}}|{\theta _d}} \right) + \mathcal{{\cal G}}\left( {{{\widehat F}^j},i|{\theta _g}} \right),\text{ }{\rm{ for }} \text{ } i \ne 0,  \label{eq:pred2}
\end{align}
where $\mathcal{G}(\cdot , j| \theta_g)$ projects the input feature to the space of $F^{j}$. 
\wh{$\hat{\cdot}$ denotes that, the feature \ly{decoded} from the bit stream would be suffering from compression loss.}
\wh{In practice, the VCM problem can be rephrased in Eq.~\eqref{eq:vcm1} by designing more task specific terms $C(\cdot | \theta_c )$, $D(\cdot | \theta_d)$ and $\mathcal{G}(\cdot| \theta_g)$.}


\subsection{A VCM Paradigm: Digital Retina}

\wht{As a VCM instance, digital retina~\cite{smart_visual_sensing,digital_Retina,knowledge_service} is to solve the problem of real-time surveillance video analysis collaboratively from massive cameras in smart cities.}
\wh{Three streams are established for human vision and machine vision as follows:}
\begin{itemize}
	\item \textbf{\wh{Video} stream}: \wht{Compressed visual data is transmitted	to learn data-driven priors to impact the optimization function in Eq.~\eqref{eq:vcm1}.
	In addition, the fully reconstructed video is optionally for humans as requested on demand.}
	\item \textbf{Model stream}: \wh{To improve the task-specific performance and hybrid video coding efficiency, hand-crafted or learning based models play a significant role. This stream is expected to guide the model updating and the model based prediction subsequently. The model learning works on the task-specific performance metric in Eq.~\eqref{eq:task_performance}.}
	\item \textbf{Feature stream}: \wh{This stream consisting of task-specific semantic or syntactic features extracted at the front end devices is transmitted to the server end. As formulated in Eq.~\eqref{eq:vcm1}, a joint optimization is expected to reduce the resource cost of video and feature streams.}
\end{itemize}

\wh{Instead of working on video stream alone, the digital retina may leverage multiple streams of features to reduce the bandwidth cost by transmitting task-specific features, and \ly{balance} the computation load by moving part of feature computing from the back end to the front end.} \wh{As an emerging VCM approach, the digital retina is revolutionizing the vision system of smart cities. However, for the current digital retina solution, the optimization in Eq.~\eqref{eq:vcm1} is reduced to minimizing an objective for each single stream separately, \whtt{without the aid of feature prediction in Eq.~\eqref{eq:pred}} and~\eqref{eq:pred2}, rather than multiple streams jointly. For example, state-of-the-art video coding standards \whtt{like HEVC} are applied to compress video streams. The compact visual descriptor standards CDVS/CDVA are applied to compress visual features. The collaboration between two different types of streams works in a combination mode. There is a lack of joint optimization across streams in terms of task-specific performance and coding efficiency.}

\wh{Generally speaking, the \whtt{traditional} video coding standardization on HEVC/AVS, together with recent compact feature descriptor standardization on CDVS/CDVA, targets efficient data exchange between human and human, machine and human, and machine and machine. To further boost the functionality of feature streams and open up more room for collaborating between streams, the digital retina introduces a novel model stream, which takes advantage of pixel-level features,  semantic features (representation), and even existing models to generate a new model for improving the performance as well as \ly{the} generality of task-related features.}
\wh{More recent work~\cite{digital_Retina} proposed to reuse multiple models to improve the performance of a target model, and further came up with a collaborative approach~\cite{digital_Retina2} to low cost and intelligent visual sensing analysis. However, how to improve the video coding efficiency and/or optimize the performance by collaborating different streams is still an open and challenging issue.}

\subsection{Reflection on VCM}

\ly{Prior to} injecting model streams, the digital retina~\cite{Digital_Retina_tian} has limitations. First, the video stream and feature stream are \ly{handled separately}, which limits the utilization of more streams.
Second, the processes of video coding and feature coding
are uni-directional, thereby limiting the optimization performance
due to the lack of feedback mechanism, which is crucial from the perspective
of a vision system in smart cities.

\wht{Beyond the basic digital retina solution, VCM is supposed to jointly optimize the compression and utilization of \ly{feature, video, and model streams} in a scalable way by introducing feedback mechanisms as shown in Fig.~\ref{fig:VCM_framework}:}

\wh{\begin{enumerate}
	\item \textbf{\textit{Feedback for collaborative mode}}: 
	\wht{The pixel and/or semantic features, equivalently video and feature streams, can be jointly optimized towards higher coding efficiency for humans and/or machines in a collaborative manner.
	That is, the features can be fed back to each other between streams for improving the task-specific performance for machines, in addition to optimizing the coding efficiency for humans.
	Advanced learning-based prediction or generation models may be applied to bridge the gap between streams.}
	\item \textbf{\textit{Feedback for scalable mode}}: 
	When bit budget cannot suffice for video or feature streams, or namely, the quality of reconstructed feature and video are not desirable, \wht{more \ly{additional resources are} utilized, along with the previously coded streams, to improve the quality of both feature and video streams with a \ly{triggered} feedback. 
	Therefore, the desired behavior of incrementally improving the performance of those human and machine vision tasks subject to an increasing bit budget can be expected.}
\end{enumerate}}

\vspace{2mm}

\begin{figure*}[t]
	\centering
	\includegraphics[width=0.8\linewidth]{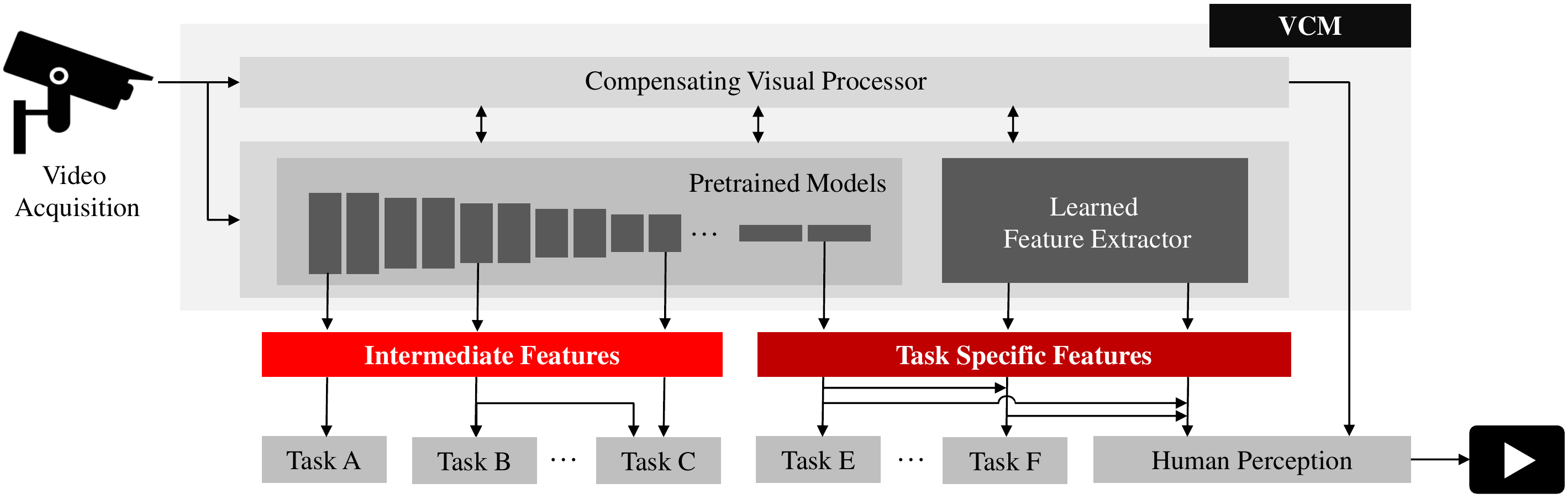}
	\caption{The key modules of VCM architecture, as well as the relationship between features and human/machine tasks.}
	\label{fig:framework4}
\end{figure*}

\section{New Trends and Technologies}
\label{sec:trend}
\wh{The efficient transition between the features of different granularities is essential for VCM. In this section, we review more recent advances in the image/video predictive and generative models.}

\subsection{Predictive Models}
\subsubsection{Image Predictive Models}
Deep convolutional neural networks~\cite{Krizhevsky_2012,residual_learning} have been proven to be effective to predict semantic labels of images/videos.
This superior capacity has been witnessed in many visual tasks, \textit{e.g.} image classification~\cite{Russakovsky_2015,residual_learning}, object detection~\cite{rgb_cvpr,ren_2017}, 
semantic segmentation~\cite{Long_2015_CVPR},
pose estimation~\cite{Kreiss_2019_CVPR}.
The key to the success of these tasks is to extract discriminative features to effectively model critical factors highly related to the final predictions.
Deep networks have hierarchical structures to extract features from low-level to high-level progressively.
The features at the deep layer are more compact and contain less redundant information, with very sparse activated regions.
As such, the image predictive model can \ly{capture} the compact and critical feature from a single image, which provides an opportunity to develop \ly{efficient} compression approaches.

\subsubsection{Video Predictive Models}

The deep networks designed for video analytics pay additional attention to modeling temporal dynamics and utilizing complex joint spatial and temporal correlations for semantic label prediction of videos.
In~\cite{KarpathyCVPR14}, several approaches extend the temporal connectivity of a CNN to fully make use of local spatio-temporal information for video classification.
In~\cite{Simonyan14b}, a two-stream ConvNet architecture incorporated with spatial appearance and motion information is built for action recognition.
\ly{Their successive works are based on} mixed architectures with both CNN and RNN~\cite{7558228}, and 3D convolutional networks~\cite{qiu2017learning} for action recognition~\cite{action_liu_2019,action_song_2018},
scene recognition~\cite{qiu2017learning}, 
captioning, commenting~\cite{Li_2016}.
These models are capable to extract discriminative and compact joint spatial and temporal features, potential to benefit squeezing out the redundancy in videos.

\subsection{Generative Models}

\subsubsection{Generative Adversarial Networks}
The \wh{advance} of generative adversarial networks (GANs)~\cite{NIPS2014_5423} \wh{makes a significant impact in \ly{machine} vision.}
\wh{Recent years have witnessed} the prosperity of image generation and its related field~\cite{2017arXiv171010196K,synthesis_gan,starGAN,TextureGAN}.
\ly{In general, most of the existing methods} can be categorized into two classes: supervised and unsupervised methods.
In supervised methods~\cite{SRGAN,SFTGAN,one_to_many}, GANs \wh{act} as a powerful loss function to capture the visual property that the traditional losses \wh{fail to} describe.
Pix2pix~\cite{pix2pix} is a milestone \wh{work} based on conditional GAN to apply the image-to-image translation from the perspective of domain transfer.
Later on, more efforts~\cite{2017arXiv171010196K,HR_image_synthesis} are \wh{dedicated to} generating high-resolution photo-realistic images with a progressive refinement framework. 
In unsupervised methods, due to the lack of the paired ground truth, the cycle reconstruction consistency~\cite{CycleGAN,DualGAN,cross_domain_gan} is introduced to model the cross-domain mapping.

\subsubsection{Guided Image Generation}
Some works focus on guided image generation, where semantic features, 
\textit{e.g.} human pose and semantic map, are taken as the guidance input.
The early attempt pays attention to pose-guided image generation and 
a two-stage network PG$^2$~\cite{NIPS2017_6644} is built to coarsely generate 
the output \wh{image} under the target pose in the first stage, and then refine it in the second stage.
In~\cite{DeformableGAN}, deformable skips are utilized to transform high-level features of each body part to better model shapes and appearances.
In~\cite{Unseen_pose}, the body part segmentation masks are used as guidance for image generation. 
However, the above-mentioned methods~\cite{NIPS2017_6644,DeformableGAN,Unseen_pose} rely on paired data. 
To \wh{address} the limitation, in~\cite{Unsupervised_synthesis}, a fully unsupervised GAN is designed, inspired by~\cite{CycleGAN, reed2016learning}.
Furthermore, the works in~\cite{Disentangled_Gen,App_shape_gen} \wh{resort to} sampling from the feature space based on the data distribution.
These techniques bring in precious opportunities to develop \ly{efficient} video coding techniques.
The semantic feature guidance is much more compact\ly{.}
\wh{With} the support of the semantic feature, the original video can be well reconstructed \wh{economically}.

\subsubsection{Video Prediction and Generation}

Another branch of generation \ly{models are for} video prediction and generation.
Video prediction \wh{aims to produce} future frames based on previous frames of a video sequence in a deterministic manner, \wh{ in which recurrent neural networks are often used to model the temporal dynamics~\cite{lstm_patch,Finn_video_prediction,2016arXiv160508104L}.}
In~\cite{lstm_patch}, an LSTM encoder-decoder network is utilized to learn patch-level video representations. 
In~\cite{Finn_video_prediction}, a convolutional LSTM is built to predict video frames.
In~\cite{2016arXiv160508104L}, a multi-layer LSTM is constructed to progressively refine the prediction \ly{process}.
Some methods do not rely on recurrent networks, \textit{e.g.} 3D convolutional neural network~\cite{2015arXiv151105440M,Futuregan}.
\wh{Some other} methods~\cite{Video_imagination,Dynamic_filter,2017arXiv170108435V} estimate local and global transforms and then apply these transforms to generate future frames indirectly.

Comparatively, video generation methods aim to produce \wh{visually} authentic video sequences in a probabilistic manner.
In the literature, methods based on GAN~\cite{Vondrick,GAN_SVC,2017arXiv171011252B,pmlr-v80-denton18a} and 
Variational AutoEncoder (VAE)~\cite{NIPS2016_6552,Frame_forecasting,2018arXiv180401523L,MoCoGAN,2017arXiv170608033V} are built.
The above-mentioned methods \wh{just predict} a few video frames.
Later works~\cite{AVP} \wh{target} long-term video prediction.
In~\cite{AVP}, up to 100 future frames of an Atari game are generated.
The future frame is generated with the guidance of the encoded features by CNN and LSTM from previous frames.
In~\cite{Villegas}, a new model \wh{is proposed to generate real video sequences.} 
The high-level structures are estimated from past frames, which are further propagated to those of the future frame via an LSTM.
In~\cite{2018arXiv180604768W}, an unsupervised method is built to extract a high-level feature which is further used to predict the future high-level feature.
After that, the model can predict future frames based on the predicted features and the first frame.

\wh{Undoubtedly, inter prediction plays a significant role in video coding. The video prediction and generation models are expected to leverage compact features to propagate the context of video for improving coding efficiency.}

\subsection{Progressive Generation and Enhancement}

Deep learning brings in the wealth of `deep'. That is, with the hierarchical layer structure, the information is processed and distilled progressively, which benefits both high-level semantic understanding and low-level vision reconstruction.
For VCM, an important property is scalability, a capacity of feature prediction ranging from the extreme compactness to the injection of redundant visual data for serving humans and machines, which is closely correlated to the hot topics of  deep progressive generation and enhancement.

A series of works have been proposed to generate or enhance images progressively.
In~\cite{2015arXiv150605751D}, a cascade of CNN  within a Laplacian pyramid framework is built to generate images in a coarse-to-fine fashion.
In~\cite{8434354}, a similar framework is applied for single-image super-resolution.
Later works refining features progressively at the feature-level, like ResNet~\cite{resNet}\ly{,} or concatenating and fusing features from different levels, like DenseNet~\cite{denseNet}, lead to better representations of pixels and their contexts for low-level visions.
The related beneficial tasks include super-resolution~\cite{2016_Yang_DEGREE,SRGAN,Zhang_2018_CVPR,MemNet,RCAN}, rain removal~\cite{jorder,DID-MDI}, dehazing~\cite{Zhang_2018_CVPR}, inpainting~\cite{dong_2018}, compression artifacts removal~\cite{gan_artifacts}, and deblurring~\cite{Kupyn_2018_CVPR}.
Zhang \textit{et al.}~\cite{Zhang_2018_CVPR} combined the structure of ResNet and DenseNet. Dense blocks are used to obtain dense local features. All features in each dense block are connected by skip connections, and then fused in the last layer adaptively in a holistic way.

In video coding, there is also a \ly{similar} tendency to pursue a progressive image and video compression, namely, \textit{scalable image/video coding~}\cite{mpeg4_scalable}, affording to form the bit-stream at any \wh{bitrate}.
The bit-stream \wh{usually consists of} several code layers (one base layer and several enhancement layers).
The base layer is responsible for \wh{basic but coarse} modeling of image\wh{/video}.
The enhancement layers progressively improve the reconstruction quality with additional bit-streams.
\wh{A} typical \wh{example} is JPEG2000~\cite{952804}, where an image pyramid from the wavelet transform is \wh{build up for scalable image reconstruction based on compact feature representations.}
Later on, \wh{the extension of the scalable video codec is made in the H.264 standard~\cite{4317636}}.
The base layer bit-stream \wh{is formed by} compressing the original \wh{video frames}, and the enhanced layer bit-stream is \wh{formed by encoding} the residue signal.

\wh{VCM is expected to incorporate the scalability into the collaborative coding of multiple task-specific data streams for humans and/or machines. It is worthy to note that,} the rapid development of DNNs (deep neural networks) 
\wh{is proliferating scalable coding schemes.}
In~\cite{8100060}, an RNN \wh{model} is utilized to realize variable-bitrate compression.
 In~\cite{2018arXiv181209443Z}, 
bidirectional ConvLSTM \wh{is} adopted to decompose the bit plane \wh{by efficiently memorizing long-term dependency.}
In~\cite{chen_zhibo_scalable}, inspired by the self-attention, a transformer-based decorrelation unit is designed to reduce the feature redundancy at different levels.
\wh{More recently}, \wh{several} works~\cite{2019arXiv191206348L,2018arXiv180409963C,8803255,xia2020emerging,hu2020coding} \wh{attempt} to jointly compress videos and features \wh{in} a scalable way, which \wh{shows more evidence for the scalability in VCM.}

\section{Video Coding for \ly{Machines}: Potential Solutions}
\label{sec:template}
In this section, 
\wh{we present several exemplar VCM \ly{solutions}:} deep intermediate feature compression,
predictive coding with collaborative feedback,
and enhancing predictive coding with scalable feedback.
\wh{Based on the fact that the pretrained \whtt{deep learning} networks, \textit{e.g.} VGG and AlexNet, can support a wide range of tasks, such as classification and object detection, we first investigate \ly{the issue of compressing} intermediate features (usually before pool5 layer) extracted from \ly{off-the-shelf} pretrained networks \whtt{(left part of Fig.~\ref{fig:framework4})}, \ly{less} specific to given tasks, via \ly{state-of-the-art} video coding techniques.
\ly{In} the next step, we explore the solution that learns to extract key points as a highly discriminative image-level feature \whtt{(right part of Fig.~\ref{fig:framework4})} to support the motion-related tasks for both machine and human vision with collaborative feedback in a feature assisted predictive way.
Finally, we \ly{attempt} to pursue  a preliminary scheme to offer a general feature \ly{scalability} to \ly{derive} both pixel and image-level representations \ly{and improve} the coding performance incrementally\ly{.}}

\begin{figure*}[t]
	\centering
	\subfigure[]{
		\includegraphics[width=0.75\linewidth]{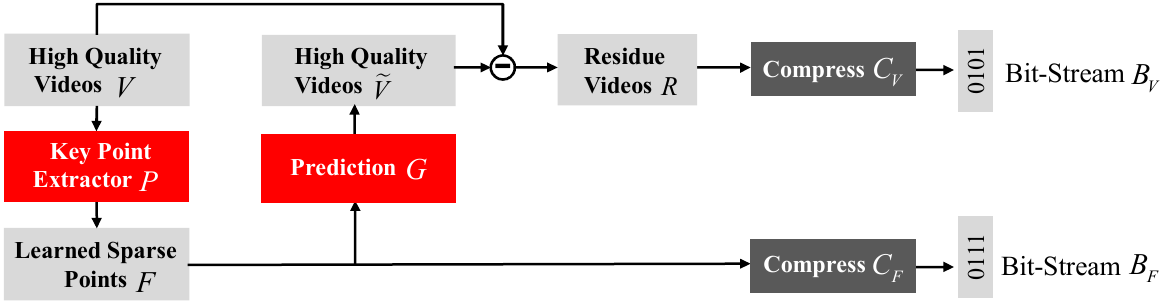}}
	\\
	\subfigure[]{
		\includegraphics[width=0.75\linewidth]{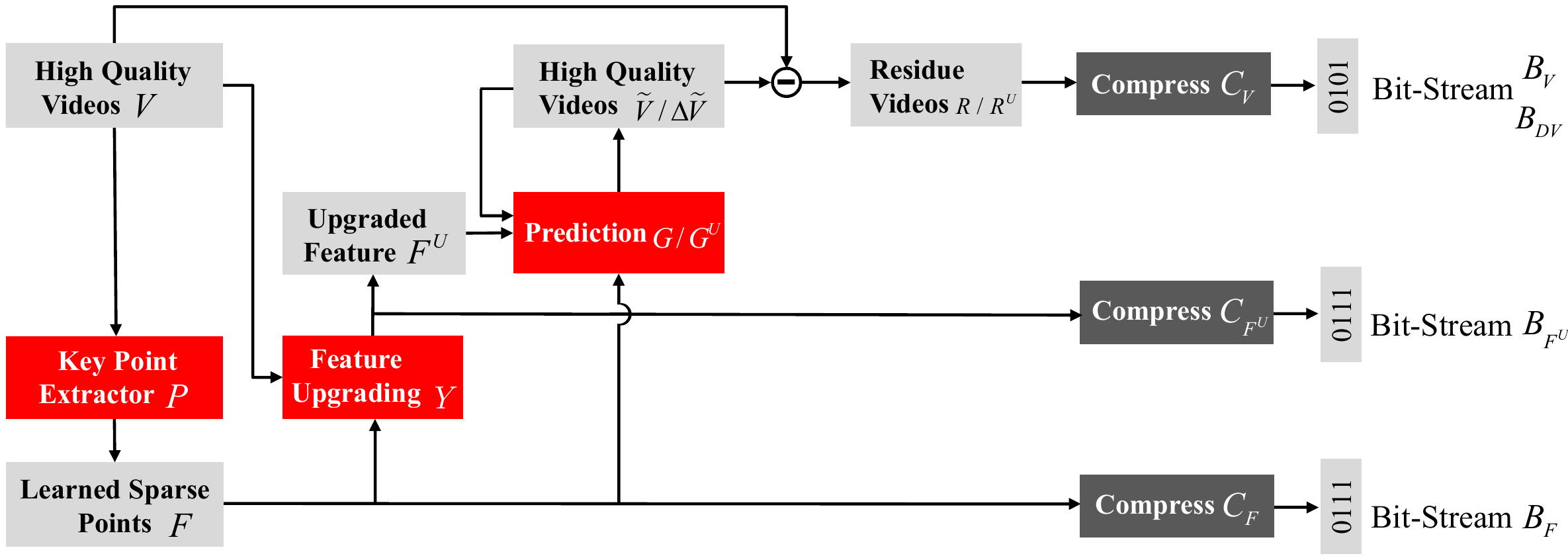}} 
	\\
	\caption{Two potential VCM solutions: (a) Predictive coding with collaborative feedback, and (b) Enhancing predictive coding with scalable feedback. }
	\label{fig:framework3}
\end{figure*}

\subsection{Deep Intermediate Feature Compression}
In the VCM, features are the key bridge for both front and back user-ends as well as high and low-level visions.
It naturally raises questions about the optimized feature extraction and model updating in the VCM framework.
For deep model-based applications, the feature compression is hindered by that, the models are generally tuned for specific
tasks, and that, the top-layer features are very task-specific and hard to be generalized.
The work in~\cite{8848858} explores a novel problem: \textit{the intermediate layer feature compression}, reducing the computing burden while being capable to support different kinds of visual analytics applications.
\ly{In practice, i}t provides a compromise between the traditional video coding and feature compression and yields a good trade-off among the computational load, communication cost, and the generalization capacity.

As illustrated in Fig.~\ref{fig:framework4}, VCM attempts to connect the features of different granularities to the human/machine vision tasks from the perspective of a general deep learning framework.
The intermediate layer features are compressed and transmitted instead of the original video \wh{or} top layer features.
Compared with the deep layer features, the intermediate features from shallow layers contain more informative cues in a general sense, as the end-to-end learning usually makes the deep layer features more task-specific with a large receptive field. To accommodate a wide range of machine vision tasks, VCM prefers to optimize the compression of intermediate features for general purposes. An interesting view is that, the shallow layers closer to the input image/video, are supposed to be less important \ly{than} the deep layers, as the semantics play a significant role in data compressing even for humans. As indicated in Fig.~\ref{fig:framework4}, VCM prefers to use the deep layers features in improving coding efficiency for humans. 

Beyond that, a further idea is to propose the problem of \textit{feature recomposition}.
\wh{The features for different tasks are} with various granularities.
It is worthwhile to explore how to evaluate the feature coding performance of all tasks in a unified perspective, \wh{and further} decide to recompose the feature of different tasks, sharing common features and organizing the related compression in a more or less scalable way.

\subsection{Predictive Coding with Collaborative Feedback}

Fig.~\ref{fig:framework3}~(a) shows the overview pipeline of a joint feature and video compression approach~\cite{xia2020emerging,hu2020coding}.
At the encoder side, a set of key frames ${{{v}}_k}$ will be first selected and compressed with traditional video codecs to form the bit-stream ${B_I}$.
The coded key frames convey the appearance information and are transmitted to the decoder side to synthesize the reconstructed non-key frames.
Then, the network learns to represent \wh{$V=\left\{ v_1, v_2, ..., v_N \right\}$} with the learned sparse points $F=\left\{ f_1, f_2, ..., f_N \right\}$ to describe temporal changes and  object motion among frames.
We \ly{employ} prediction model $P\left( \cdot \right)$ and generation model $G\left( \cdot \right)$ to implement the feature transition operation $\mathcal{G}(\cdot)$ to convert the feature from a redundant \ly{form} to compact one and vice verse, respectively.
More specifically, $P(\cdot)$ and $G(\cdot)$ are the processes to extract key points from videos and generate videos based on the learned key points. To extract the key points, we have:
\begin{equation}
F = P\left( {V,\lambda |{\theta _{gp}}} \right),
\end{equation}
where ${\theta _{gp}}$ is a learnable parameter. $F$ is a compact feature, which only requires very fewer bit streams for transmission and storage. 
$\lambda $ is a rate control parameter. 
The compression model ${{{C}}_F}\left( { \cdot |{\theta _{cf}}} \right)$ compresses $F$ into the feature stream ${B_F}$:
\begin{equation}
{B_F}{\rm{ = }}{{{C}}_F}\left( {F|{\theta _{cf}}} \right),
\end{equation}
where ${\theta _{cf}}$ is a learnable parameter.

Then, a motion guided generation network \wh{calculates} the motion based on these points and then transfers the appearance from the reconstructed key frames to \ly{those} remaining non-key frames.
Specifically, for the $t$-th frame to be reconstructed, we \ly{denote} its previous reconstructed key frame, previous reconstructed key points, and the current reconstructed key points \ly{by} ${{{\widehat v}_{\psi(t)}}}$,${{\widehat{f}_{\psi(t)}}}$  and ${{\widehat{f}_{t}}}$:
 where $\psi(t)$ maps the index of the key frame of the $t$-th frame.
The target frame ${{\widetilde v}_t} \in \widetilde{V}$ is synthesized as follows:
\begin{equation}
{{\widetilde v}_t} = G \left( {{{\widehat v}_{\psi(t)}},{{\widehat f}_{\psi(t)}},{{\widehat f}_t}} | \theta_{gg} \right),
\end{equation}
where ${\theta _{gg}}$ is a learnable parameter.
After that, the residual video $R=V-\widetilde{V}$ can be calculated, where $\widetilde{V}=\left\{ \widetilde{v}_1, \widetilde{v}_2, ..., \widetilde{v}_N \right\}$.
The sparse point descriptor and residual video will be quantized and compressed to the bit stream ${B_F}$ and 
${B_V}$ for transmission. That is ${B_V} = \left\{B_I, B_R \right\}$. We can \ly{adjust} the total bitrate via controlling the bitrates of ${B_F}$ and ${B_V}$.

At the decoder side, the key frames will be first reconstructed from ${{{B}}_{{I}}}$.
The sparse point representations are also decompressed as ${{\widehat F}} = \left\{ {{{\widehat f}_1},{{\widehat f}_2},...,{{\widehat f}_N}} \right\}$ from $B_F$ as follows,
\begin{equation}
\widehat F{\rm{ = }}{{{D}}_F}\left( {{B_F}|{\theta _{df}}} \right),
\end{equation}
where ${{{D}}_F}\left( { \cdot |{\theta _{df}}} \right)$ is a feature decompression model, and ${\theta _{df}}$ is a learnable parameter.
The videos are reconstructed via: 
$\widehat{V} = \widetilde{V} +\widehat{R} $.
Finally, $\widehat F$ along with $\widehat{V}$ 
serves machine analysis and human vision, respectively.

\subsection{Enhancing Predictive Coding with Scalable Feedback}

Fig.~\ref{fig:framework3}~(b) shows an optimized architecture for VCM, \wh{by adding} scalable feedback. Similarly, the key points of video frames $F$ are extracted.
After that, the redundancy of the video and key frames ${{{\widehat v}_{\psi(t)}}}$ is removed by applying feature-guided prediction. 
Then, the residue video $R$ is compressed into the video stream, which is further passed to the decoder.

When the feature and video qualities after decompression do not meet the requirements, a scalable feedback is launched and more guidance features are introduced:
\begin{align}
\Delta F & = Y\left( F,  V  | \theta_y \right), \\
F^U & = F + \Delta F,
\end{align} 
where $\theta_y$ is a learnable parameter.
With the key points $F$ in the base layer (the result in the scheme in Fig.~\ref{fig:framework3}~(a)) and the original video $V$, we generate the residual feature $\Delta F$ to effectively utilize additionally allocated bits to encode more abundant information and form the updated feature $F^U$.

Then, $G^U$ is used  to refine $V$ by taking the reconstructed key points and video as its input:
\begin{equation}
\Delta \widetilde{V} = G^U\left( \widehat{F}, \Delta \widehat{F}, \widehat V  | \theta_h \right),
\end{equation}
where $\theta_h$ is a learnable parameter.
Then, we can infer the incremental residue video: ${R^U} = V - (\widetilde V + \Delta \widetilde V) - R$, which is \ly{compressed} into the bit stream $B_{DV}$.
At the decoder side, \wh{${\hat{R}^U}$} is decompressed from $B_{DV}$.
Then, the video with a high quality is inferred via: \wh{${\widehat V^U} = \widetilde V + \Delta \widetilde V + {\widehat{R}^U} + \widehat{R}$}.
This \ly{allows} to introduce more bits via launching scalable feedback.

\vspace{2mm}

\section{Preliminary Experimental Results}
\label{sec:results}
In this section, we provide preliminary experimental results \ly{from} the perspectives of intermediate deep feature compression and machine-human collaborative compression.

\begin{table}[]
	\scriptsize
	\caption{Lossy feature compression results for different tasks (Comp.Rate|Fidelity$^2$)}
	\label{tab:retrival}	
	\setlength{\tabcolsep}{0.6mm}{
	\begin{tabular}{c|cccc|cccc|cccc}
		\hline
		Feature & \multicolumn{4}{c|}{Classification} & \multicolumn{4}{c|}{Retrieval} & \multicolumn{4}{c}{Detection} \\
		\hline
		QP      & \multicolumn{2}{c}{22} & \multicolumn{2}{c|}{42} & \multicolumn{2}{c}{22} & \multicolumn{2}{c|}{42} & \multicolumn{2}{c}{22} & \multicolumn{2}{c}{42} \\	
		\hline
		\multicolumn{13}{c}{VGGNet}                                                                                 \\
		\hline
		conv1   & 0.080      & 0.985     & 0.020      & 0.839     & 0.041      & 0.997     & 0.006      & 0.955     & 0.065      & 0.954     & 0.013      & 0.850     \\
		pool1   & 0.099      & 0.984     & 0.023      & 0.693     & 0.039      & 0.996     & 0.005      & 0.923     & 0.085      & 0.942     & 0.018      & 0.820     \\
		conv2   & 0.098      & 0.972     & 0.035      & 0.790     & 0.069      & 0.996     & 0.027      & 0.955     & 0.090      & 0.950     & 0.030      & 0.858     \\
		pool2   & 0.138      & 0.982     & 0.047      & 0.745     & 0.119      & 0.997     & 0.037      & 0.945     & 0.128      & 0.953     & 0.040      & 0.815     \\
		conv3   & 0.080      & 0.986     & 0.034      & 0.840     & 0.089      & 0.998     & 0.048      & 0.976     & 0.033      & 0.954     & 0.015      & 0.845     \\
		pool3   & 0.140      & 0.981     & 0.063      & 0.819     & -          & -         & -          & -         & 0.066      & 0.955     & 0.032      & 0.826     \\
		conv4   & 0.053      & 0.984     & 0.028      & 0.865     & 0.070      & 0.997     & 0.043      & 0.959     & 0.019      & 0.960     & 0.008      & 0.877     \\
		pool4   & 0.127      & 0.974     & 0.065      & 0.864     & -          & -         & -          & -         & 0.041      & 0.960     & 0.019      & 0.847     \\
		conv5   & 0.046      & 0.989     & 0.023      & 0.920     & 0.041      & 0.995     & 0.030      & 0.952     & 0.023      & 0.956     & 0.005      & 0.741     \\
		pool5   & 0.129      & 0.986     & 0.075      & 0.908     & 0.200      & 0.996     & 0.146      & 0.960     & -          & -         & -          & -         \\
		\hline
		\multicolumn{13}{c}{ResNet}                                                                                  \\
		\hline
		conv1   & 0.041      & 0.935     & 0.005      & 0.356     & 0.018      & 0.964     & 0.001      & 0.792     & 0.029      & 0.915     & 0.002      & 0.713     \\
		pool1   & 0.043      & 0.937     & 0.004      & 0.087     & 0.025      & 0.963     & 0.002      & 0.720     & 0.030      & 0.889     & 0.002      & 0.470     \\
		conv2   & 0.095      & 0.986     & 0.014      & 0.765     & 0.027      & 0.939     & 0.001      & 0.554     & 0.107      & 0.949     & 0.016      & 0.660     \\
		conv3   & 0.134      & 0.989     & 0.028      & 0.854     & 0.041      & 0.986     & 0.003      & 0.551     & 0.118      & 0.971     & 0.021      & 0.684     \\
		conv4   & 0.170      & 0.992     & 0.034      & 0.932     & 0.035      & 0.997     & 0.012      & 0.799     & 0.056      & 0.964     & 0.008      & 0.833     \\
		conv5   & 0.131      & 0.998     & 0.063      & 0.961     & 0.040      & 0.999     & 0.014      & 0.992     & -          & -         & -          & -     \\		
		\hline
	\end{tabular}}
\end{table}

\begin{table}[]
	\scriptsize
	\caption{Lossy feature compression results for different tasks. DD-Channel Concatenation denotes channel concatenation by descending difference. (Comp.Rate|Fidelity$^2$) }
	\label{tab:exp_feature_order}	
	\setlength{\tabcolsep}{0.6mm}{	
	\begin{tabular}{c|cccc|cccc|cccc}
		\hline
		Feature & \multicolumn{4}{c|}{Channel Concatenation}       & \multicolumn{4}{c|}{DD-Channel Concatenation}    & \multicolumn{4}{c}{Channel Tiling}              \\
		\hline
		QP      & \multicolumn{2}{c}{12} & \multicolumn{2}{c|}{42} & \multicolumn{2}{c}{12} & \multicolumn{2}{c|}{42} & \multicolumn{2}{c}{12} & \multicolumn{2}{c}{42} \\
		\hline
		conv1   & 0.100       & 0.991    & 0.006       & 0.666    & 0.100       & 0.992    & 0.006       & 0.694    & 0.116       & 0.998    & 0.020       & 0.839    \\
		pool1   & 0.124       & 0.987    & 0.006       & 0.401    & 0.124       & 0.986    & 0.005       & 0.412    & 0.145       & 0.993    & 0.023       & 0.686    \\
		conv2   & 0.116       & 0.989    & 0.012       & 0.343    & 0.116       & 0.988    & 0.012       & 0.455    & 0.130       & 0.992    & 0.035       & 0.781    \\
		pool2   & 0.166       & 0.991    & 0.014       & 0.260    & 0.165       & 0.991    & 0.014       & 0.369    & 0.184       & 0.996    & 0.045       & 0.756    \\
		conv3   & 0.093       & 0.989    & 0.013       & 0.586    & 0.093       & 0.990    & 0.013       & 0.617    & 0.100       & 0.995    & 0.033       & 0.835    \\
		pool3   & 0.164       & 0.990    & 0.023       & 0.477    & 0.164       & 0.994    & 0.023       & 0.541    & 0.163       & 0.992    & 0.053       & 0.791    \\
		conv4   & 0.059       & 0.992    & 0.012       & 0.700    & 0.059       & 0.988    & 0.012       & 0.706    & 0.051       & 0.993    & 0.021       & 0.857    \\
		pool4   & 0.140       & 0.990    & 0.030       & 0.624    & 0.142       & 0.991    & 0.028       & 0.611    & 0.100       & 0.993    & 0.042       & 0.855    \\
		conv5   & 0.046       & 0.995    & 0.016       & 0.809    & 0.046       & 0.990    & 0.015       & 0.812    & 0.028       & 0.996    & 0.010       & 0.920    \\
		pool5   & 0.127       & 0.992    & 0.055       & 0.776    & 0.127       & 0.994    & 0.053       & 0.746    & 0.054       & 0.996    & 0.024       & 0.903    \\
		\hline
	\end{tabular}}
\end{table}

\subsection{Deep Intermediate Feature Compression for Different Tasks}

\subsubsection{Compression Results}
\wh{We show the compression performance on the intermediate features for different tasks in Table~\ref{tab:retrival}.}
The compression rate is calculated by the ratio of original intermediate deep features and the compressed bit-streams.
As to the fidelity evaluation\footnote{https://github.com/ZoomChen/DeepFeatureCoding/tree/master/Coding\_and\\\_Evaluation}, the reconstructed features are passed to their birth-layer of the
original neural network to infer the network outputs, which will be compared with pristine outputs to evaluate the information loss of the lossy compression methods.
More \wh{results and} details on the evaluation framework can be found in~\cite{chenzhuo_MM}.

\wh{From Table~\ref{tab:retrival}}, several interesting observations are reached.
First, the potential of lossy compression is inspiring. In the extreme case, for example in image retrieval, 
\wht{ResNet conv2 feature \wh{achieves} at least 1000$\times$ compression ratio at QP 42,}
while the lossless methods usually provide 2-5$\times$.
Second, for each feature type, the fidelity metric decreases with a larger QP value.
Third, QP 22 \ly{generally provides} high fidelity and fair compression ratio. 
Forth, \ly{upper} layer features, like conv4 to pool5, tend to be more robust to heavy compression.

\subsubsection{Channel Packaging}
Deep features have multiple channels. 
It needs to arrange these features into single-channel or three-channel maps and then compress them with the existing video codecs.
Three modes are \ly{studied}: \textit{channel concatenation}, 
\textit{channel concatenation by descending difference},
and \textit{channel tiling}.
For \textit{channel concatenation}, each channel of the feature map corresponds to a frame in the input data of a traditional video encoder. The height and width of the feature map are filled to the height and width that meet the input requirements of the video encoder. The feature map channel order is the original order and remains unchanged.
In this mode, inter-coding of HEVC is applied.
For \textit{channel concatenation by descending difference}, to obtain higher compression efficiency, the channel of the feature map  is reordered before being \ly{fed into} a traditional video encoder.
The first channel is fixed, and the remaining channels are arranged according to the L2 norm of the different between the current channel to the previous one.
For \textit{channel tiling}, multiple channels are tiled into a \ly{two-dimensional map}, serving as an \ly{input} to a video encoder.
The result is presented in Table~\ref{tab:exp_feature_order}.
These results are preliminary and \ly{more} efforts are \ly{expected to improve} the efficiency of compressing deep feature maps.

\vspace{2mm}

\subsection{Joint Compression of Feature and Video}

\wh{Let us evaluate the effectiveness of the potential solution: feature assisted predictive coding with the collaborative feedback.
We show the results of \textit{compression for machine vision},
including 
\textit{action recognition},
\textit{human detection}
and 
\textit{compression for human vision},
video reconstruction.}

\subsubsection{Experimental Settings}
PKU-MMD dataset \cite{pkummd} is used to generate the testing samples.
In total, \wht{3317 clips, each sampling 32 frames, are used for training}, and $227$ clips, each sampling $32$ frames, are used for testing. All frames are cropped and resized to $512\times512$\ly{.} To verify the coding efficiency, we use HEVC, \wht{implemented in FFmpeg version 2.8.15}\footnote{https://www.ffmpeg.org/}, as the anchor for comparison by compressing all frames with the HEVC codec in the constant rate factor mode.

\ly{To} evaluate the performance of feature assisted predictive coding, the sparse motion pattern~\cite{Siarohin_2019_CVPR} is extracted to serve machine vision.
\wht{For a given input frame, a U-Net followed by softmax activations is used to extract heatmaps for key point prediction.
The covariance matrix is additionally generated to capture the correlations between the key points and its neighbor pixels. For each key point, in total 6 float numbers including two numbers indicating the position and 4 numbers in the covariance matrix are used for description.}
The selected key frames are compressed and transmitted, along with the sparse motion pattern to generate the full picture for human vision.
In the testing process, we select the first frame in each clip as the \ly{key frames}. At the encoder side, the key frame is coded with the HEVC codec in the constant rate factor mode.
Besides the key frame, 20 key points of each frame form the sparse motion representation. Each key point contains 6 float numbers. For the two position numbers, a quantization with the step \ly{of} $2$ is performed for compression.
For the other 4 float numbers belonging to the covariance matrix, we calculate the inverse of the matrix in advance, and then quantize the 4 values with a step \ly{of} $64$.
Then, the quantized key point values are further losslessly compressed by the Lempel Ziv Markov chain algorithm (LZMA) algorithm \cite{lzma}.

\begin{table}[t]
	\centering
	\scriptsize
	\begin{minipage}{0.47\linewidth}
		\caption{Action recognition accuracy of different methods and corresponding bitrate costs.}		
		\label{tab1}
		\begin{tabular}{c|c|c}
			\hline
			Codec     & Bitrate (Kbps) & Accu.(\%) \bigstrut\\
			\hline
			\hline
			HEVC & 16.2  & 65.2  \bigstrut\\
			Ours & 5.2   & 74.6  \bigstrut\\
			\hline
		\end{tabular}
	\end{minipage}
	\hspace{2mm}
	\begin{minipage}{0.47\linewidth}
		\caption{SSIM comparison between different methods and corresponding bitrate costs.}
		\label{tab2}
		\begin{tabular}{c|c|c}
			\hline
			Codec & Bitrate (Kbps) & SSIM \bigstrut\\
			\hline
			\hline
			HEVC  & 33.0  & 0.9008  \bigstrut\\
			Ours  & 32.1  & 0.9071  \bigstrut\\
			\hline
		\end{tabular}
	\end{minipage}
\end{table}

\begin{figure}[t]
	\centering
	\subfigure{
		\includegraphics[width=0.3\linewidth]{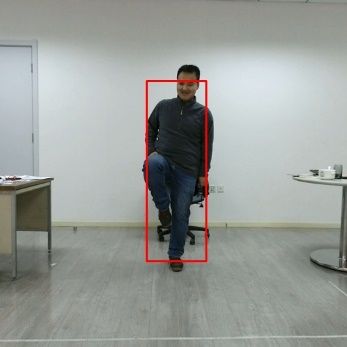}
	} \hspace{-2mm}
	\subfigure{
		\includegraphics[width=0.3\linewidth]{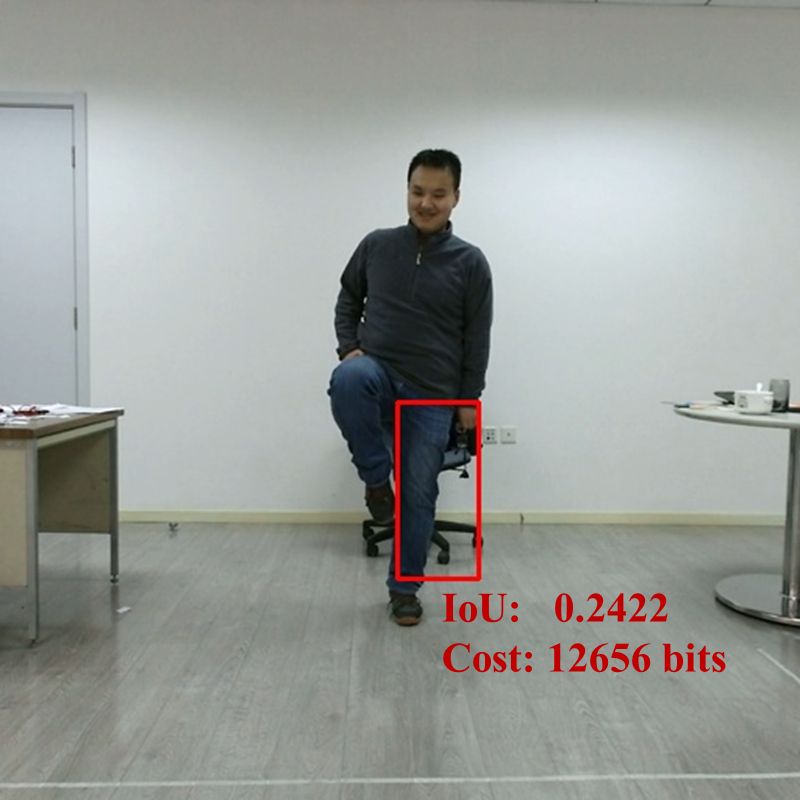}
	} \hspace{-2mm}
	\subfigure{
		\includegraphics[width=0.3\linewidth]{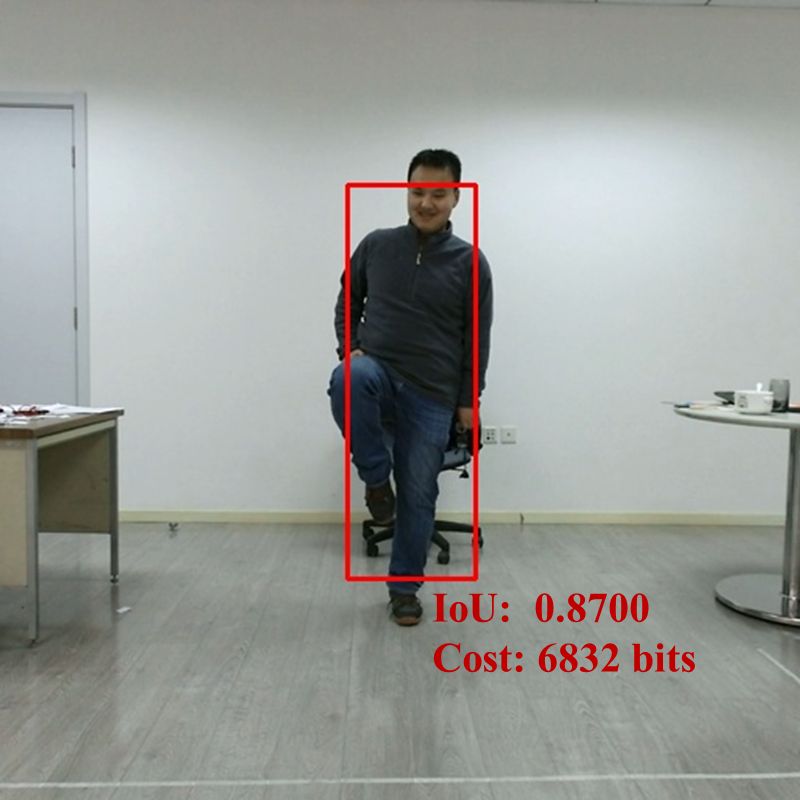}
	}
	\\
		\subfigure{
			\includegraphics[width=0.3\linewidth]{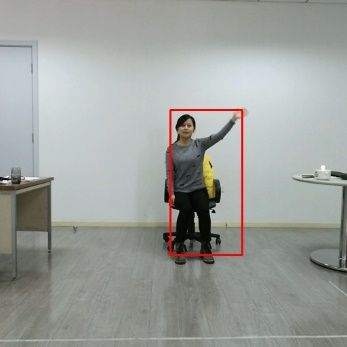}
		} \hspace{-2mm}
		\subfigure{
			\includegraphics[width=0.3\linewidth]{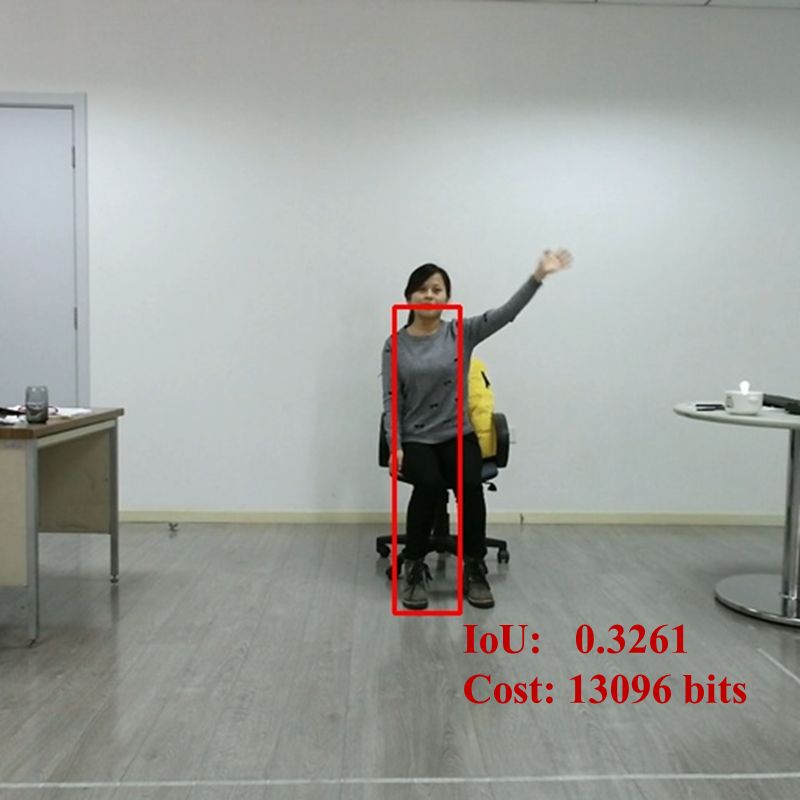}
		} \hspace{-2mm}
		\subfigure{
			\includegraphics[width=0.3\linewidth]{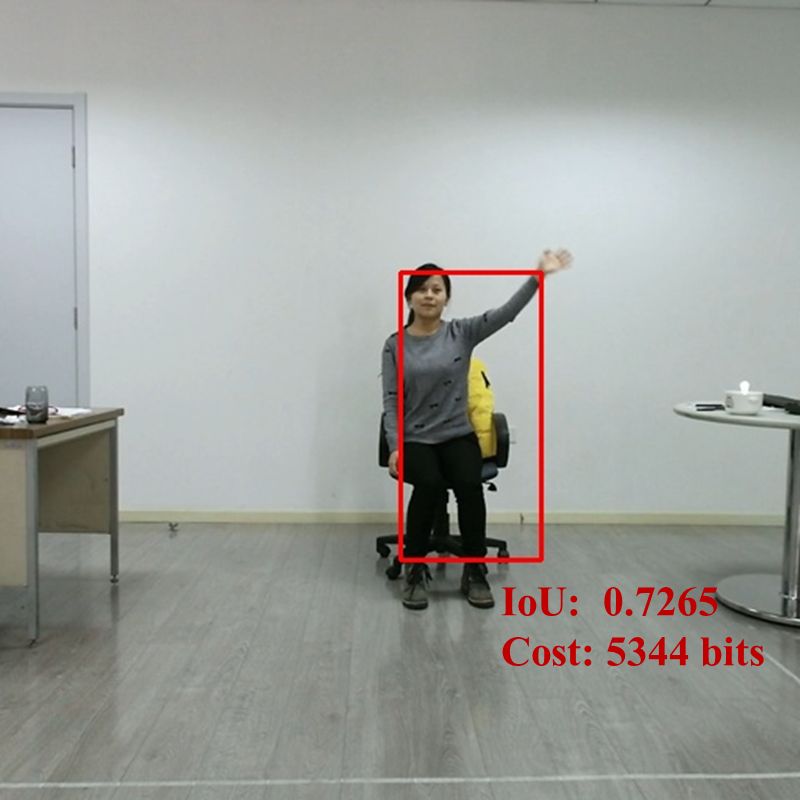}
		}
		\addtocounter{subfigure}{-6}
		\subfigure[GT]{
			\includegraphics[width=0.3\linewidth]{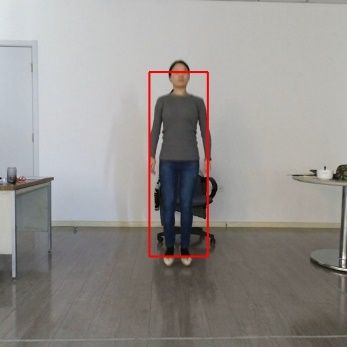}
		} \hspace{-2mm}
		\subfigure[YOLO v3]{
			\includegraphics[width=0.3\linewidth]{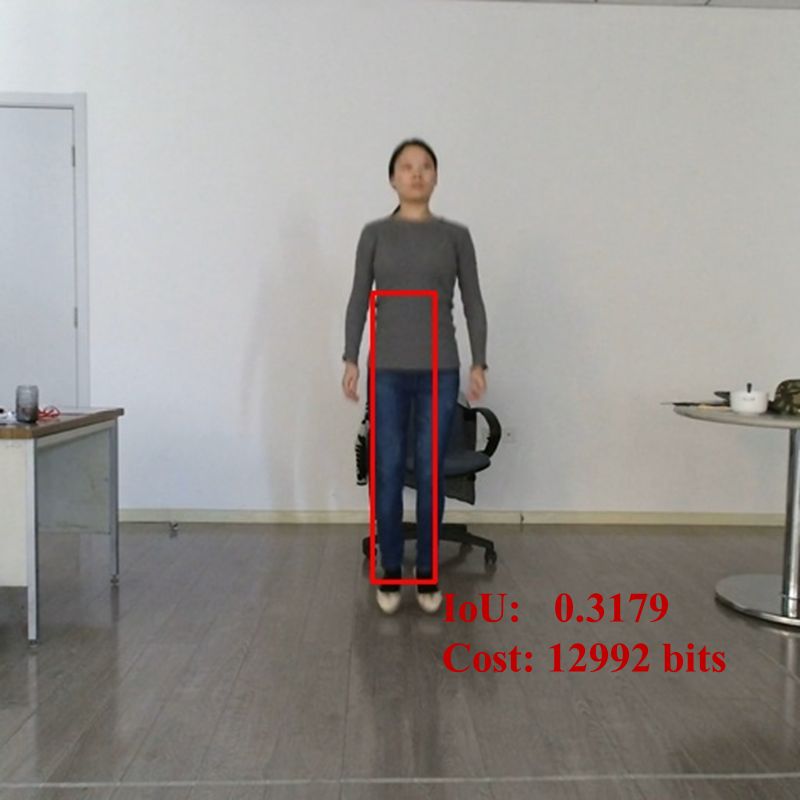}
		} \hspace{-2mm}
		\subfigure[Proposed]{
			\includegraphics[width=0.3\linewidth]{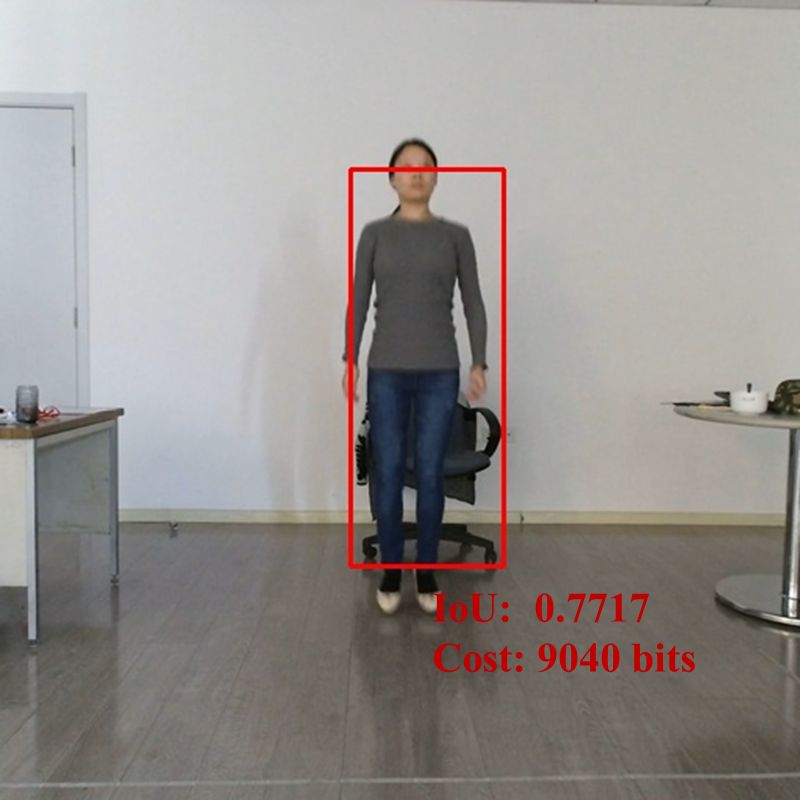}
		}
	\caption{Subjective results of human detection. The coding cost represents \ly{the} bits \ly{required} to code the corresponding testing clip.}
	\label{fig:fiou}
\end{figure}

\subsubsection{Action Recognition}
\label{actionacc}
We first \wh{evaluate} the efficiency of the learned key points for \wh{action recognition}.
Although \wh{each key point is represented with} 6 numbers, we only use two quantized position numbers for action recognition. 
To align with the bitrate cost of the features, the clips are \wh{first} down-scaled to $256\times256$ and then compressed with the constant rate factor $51$ with HEVC. \wht{All 227 clips are used in the testing.}
Table \ref{tab1} has shown the action recognition accuracy and corresponding bitrate costs of different kinds of data. Our method can obtain considerable action recognition accuracy with only $5.2$ Kbps bitrate cost, \wh{superior to that by HEVC.}

\begin{figure}[t]
	\centering
	\scriptsize
	\begin{minipage}{0.48\linewidth}
		\includegraphics[width=0.8\linewidth]{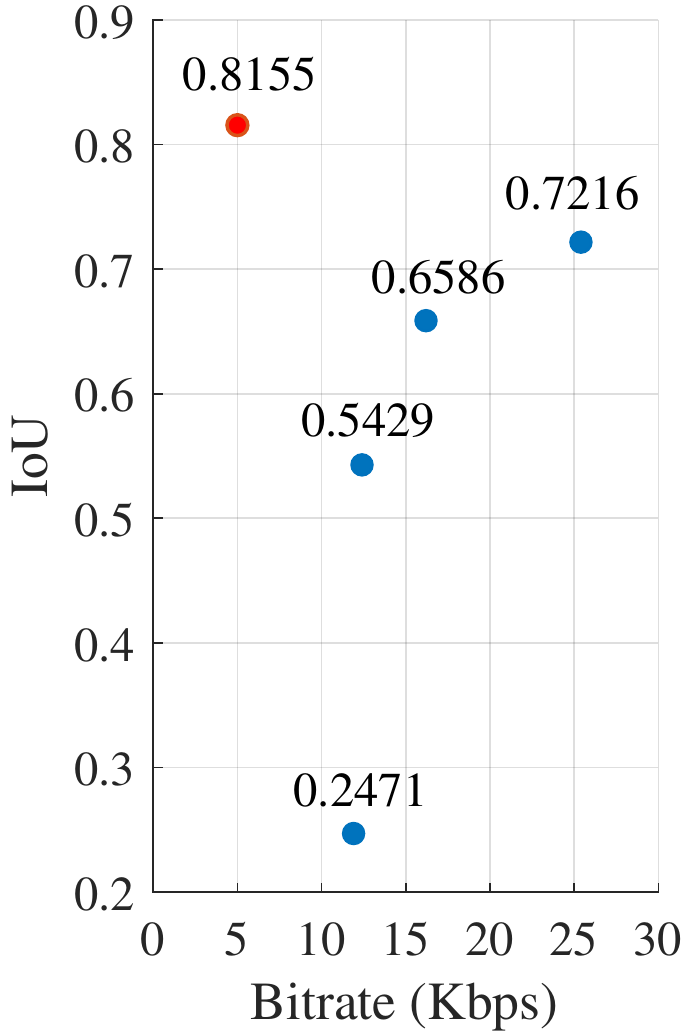}
		\caption{Human detection IoU under different \ly{bitrates}.}
		\label{fig:san}
	\end{minipage}
	\hspace{1mm}
	\begin{minipage}{0.48\linewidth}
		\includegraphics[width=0.8\linewidth]{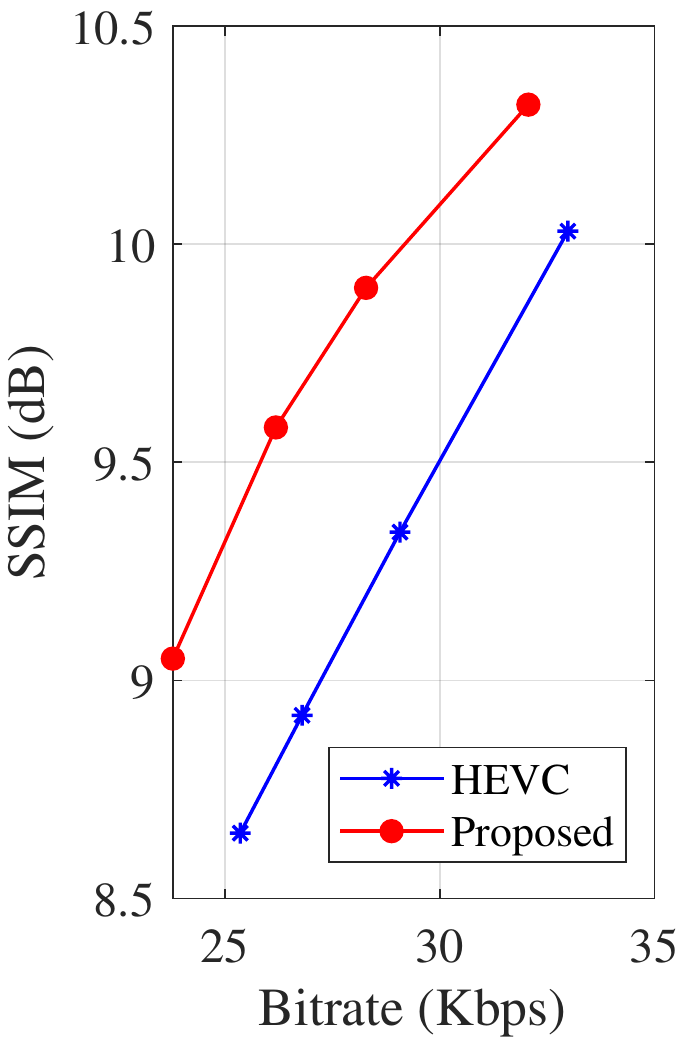}
		\caption{\wh{Rate distortion curves of HEVC and the proposed method.}}
		\label{fig:frd}
	\end{minipage}
\end{figure}


\subsubsection{Human Detection}
Apart from action recognition, human detection accuracy is also compared. The original skeleton information in the dataset is used to form the ground truth bounding box and a classical detection algorithm YOLO v3~\cite{yolo} is adopted for comparison.
All 227 clips are used in the testing.
The testing clips are down-scaled to different scales of $64\times64$, $128\times128$, $256\times256$ and $512\times512$ and compressed by HEVC with the constant rate factor $51$ to form the input of YOLO v3. Fig. \ref{fig:san} has shown the Intersection over Union (IoU) of different methods and their corresponding bitrates. Our method can achieve much better detection accuracy with lower bitrate costs. Some subjective results of human detection are shown in Fig. \ref{fig:fiou}, we can see that our method can achieve better detection accuracy with fewer bit costs.

\begin{figure}[t]
	\centering
	\begin{minipage}{0.23\textwidth}
		\centering
		\ifx \useAnimate \undefined
		\includegraphics[width=0.98\textwidth]{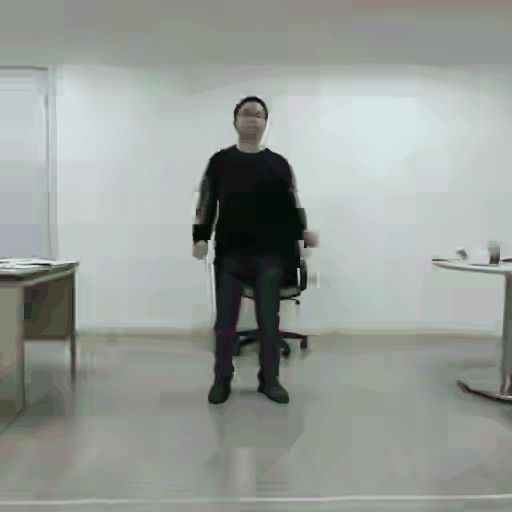}
		\else
		\animategraphics[width=0.98\textwidth,autoplay,loop]{12}{imgs/hevc/0410bd6d0e154085935147895b2254a6-}{5}{20}
		\fi
	\end{minipage}	
	\begin{minipage}{0.23\textwidth}
		\centering
		\ifx \useAnimate \undefined
		\includegraphics[width=0.98\textwidth]{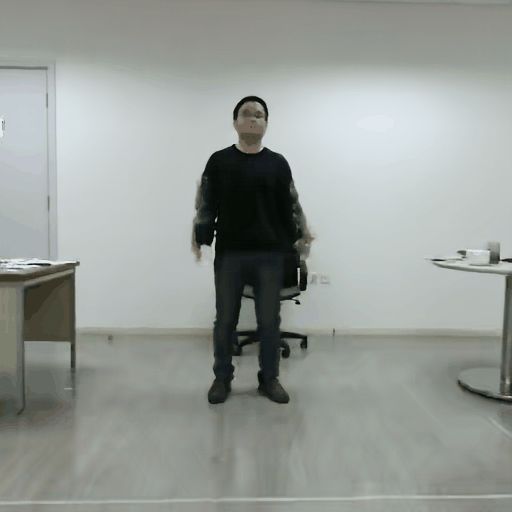}
		\else
		\animategraphics[width=0.98\textwidth,autoplay,loop]{12}{imgs/res/784dc6e3b93f4edba593639453b89297-}{5}{20}
		\fi
	\end{minipage}
	\caption{
		The visual results of the reconstructed videos by HEVC (left panel) and our method (right panel)\ly{, respectively}.
	}
	\label{fig:teaser}
\end{figure}

\begin{figure}[t]
	\centering
	\subfigure{
	\includegraphics[width=0.31\linewidth]{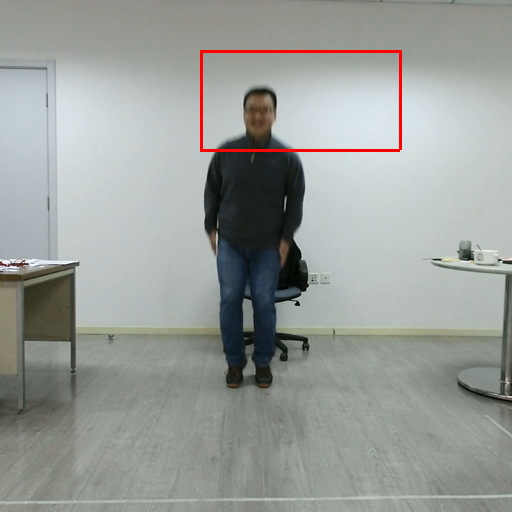}}
	\hspace{-2mm}
	\subfigure{
		\includegraphics[width=0.31\linewidth]{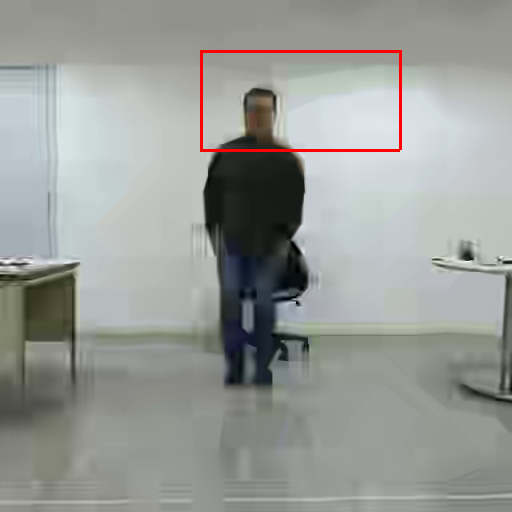}}
	\hspace{-2mm}
	\subfigure{
		\includegraphics[width=0.31\linewidth]{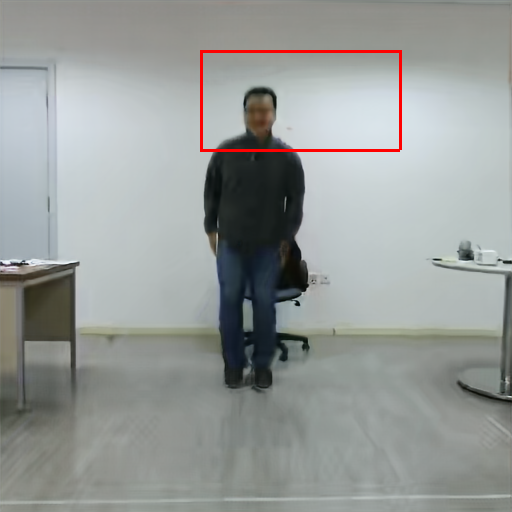}
	}
	\\ 
	\vspace{-2mm}
	\subfigure{
		\includegraphics[width=0.31\linewidth]{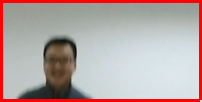}
	}
	\hspace{-3mm}
	\subfigure{
		\includegraphics[width=0.31\linewidth]{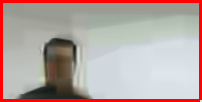}
	}
	\hspace{-3mm}
	\subfigure{
		\includegraphics[width=0.31\linewidth]{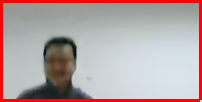}
	}
	\\	
	\subfigure{
		\includegraphics[width=0.31\linewidth]{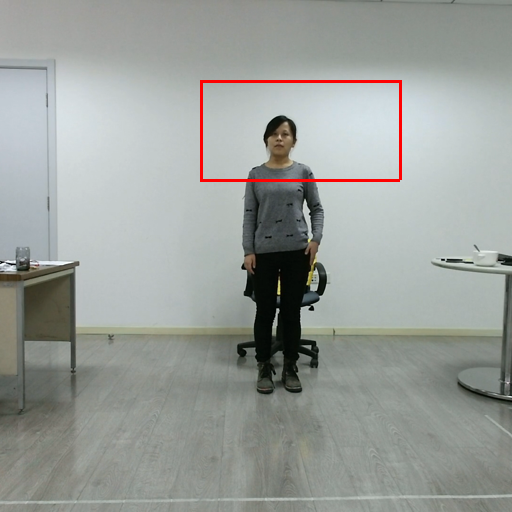}}
	\hspace{-2mm}
	\subfigure{
		\includegraphics[width=0.31\linewidth]{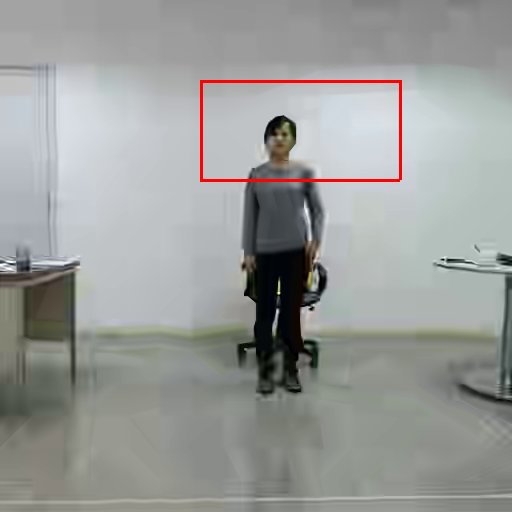}}
	\hspace{-2mm}
	\subfigure{
		\includegraphics[width=0.31\linewidth]{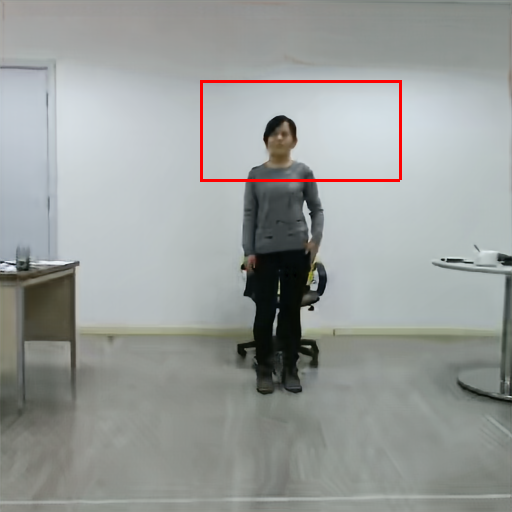}
	}
	\\ 
	\vspace{-2mm}
	\setcounter{subfigure}{0}
	\subfigure[Ground Truth]{
		\includegraphics[width=0.31\linewidth]{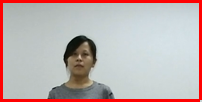}
	}
	\hspace{-3mm}
	\subfigure[HEVC]{
		\includegraphics[width=0.31\linewidth]{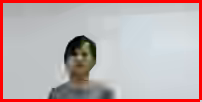}
	}
	\hspace{-3mm}
	\subfigure[Proposed]{
		\includegraphics[width=0.31\linewidth]{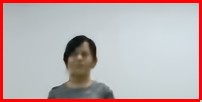}
	}
	\\ \vspace{-2mm}
	\caption{Video reconstruction results of different methods.	}
	\label{fig:f3}
	\vspace{-3mm}
\end{figure}

\subsubsection{Video Reconstruction}
The video reconstruction quality of the proposed method is \ly{compared} with that of HEVC. During the testing phase, we compress the key frames with a constant rate factor $32$ to maintain a high appearance quality.
The bitrate is calculated by \ly{considering} the compressed key frames and key points.
As for HEVC, we compress all frames with a constant rate factor $44$ to achieve an approaching bitrate\ly{.}

Table \ref{tab2} has shown the quantitative reconstruction quality of different methods. SSIM values are adopted for quantitative comparison. It can be observed that, our method can achieve better reconstruction quality than HEVC with lower bitrate cost. The subjective results of different methods are shown in Fig. \ref{fig:f3}. There are obvious compression artifacts on the reconstruction results of HEVC, which heavily degrade the visual quality. Compared with HEVC, our method can provide far more visually pleasing results.

Moreover, we add a rate distortion curve for comparison. HEVC is used as the anchor \ly{undergoing} four constant rate factors 44, 47, 49 and 51. For our \ly{method}, the key frames are compressed respectively under constant rate factors 32, 35, 37 and 40. The rate distortion curve is shown in Fig. \ref{fig:frd}. Our method \ly{yields better} reconstruction quality \ly{at} all bitrates.

\section{Discussion and Future Directions}
\label{sec:discussions}

\subsection{Entropy Bounds for Tasks}
Machine vision tasks rely on features at different levels of granularity.
High-level tasks prefer more discriminative and compact features\ly{,} while low-level tasks need abundant \ly{pixels} for fine modeling. As VCM is to explore the collaborative compression and intelligent analytics over multiple tasks, it is valuable to figure out the intrinsic relationship among a variety of tasks in a general sense.
There's some preliminary work to reveal the connection between typical vision tasks~\cite{Zamir_2018_CVPR}.
However, there is no theoretical evidence to measure the information associated with any given task. We may resort to extensive experiments on the optimal compression ratio vs. the desired performance for each specific task. But the empirical study hinders the collaborative optimization across multiple tasks due to the complex objective and the heavy computational cost. Moreover, the proposed VCM solution \ly{benefits} from the incorporation of collaborative and scalable feedback over tasks. How to mathematically formulate the connection of different tasks helps to pave the path to the completeness in theory on the feedback mechanism in VCM. In particular, the theoretical study on entropy bounds for tasks is important for VCM to improve the performance and efficiency for machine and human vision in a broad range of AI applications.

\subsection{Bio-Inspired Data Acquisition and Coding}
Recently, inspired by the biological mechanism of human vision, 
researchers invent the bio-inspired spike camera to continuously accumulate luminance intensity and launch spikes when reaching the dispatch threshold. The spike camera brings about a new  capacity of capturing the fast-moving scene in a frame-free manner while reconstructing full texture, which provides new insights into the gap between human vision and machine vision, and new opportunities for addressing the fundamental scientific and technical issues in video coding for machines. Valuable works have been done to investigate the spike firing mechanism~\cite{spike1}, spatio-temporal distribution of the spikes~\cite{spike2}, and lossy spike coding framework for efficient spike data coding~\cite{spike3}. The advance of the spike coding shows other potentials for VCM.

\subsection{Domain Shift in Prediction and Generation}
By employing the data-driven methods, the VCM makes more compact and informative features. The risk is that those methods might be trapped in the over-fitting due to the domain shift problem. Targeting more reliable VCM, how to improve the domain generalization of the prediction and generation models, and how to realize the domain adaptation (say, via online learning) are important topics.

\section{Conclusion}
\label{sec:conclusion}

As a response to the emerging MPEG standardization efforts VCM, this paper formulates a new problem of video coding for machines, targeting the collaborative optimization of video and feature coding for human and/or machine visions.
Potential solutions, preliminary results, and future direction of VCM are presented. The state-of-the-art video coding, feature coding, and general learning approaches from the perspective of predictive and generative models, are reviewed comprehensively as well. As an initial attempt in identifying the roles and principles of VCM, this work is	 expected to call for more evidence of VCM from \ly{both} academia and industry, especially when AI meets the big data era.

	\scriptsize
	\bibliographystyle{IEEEtran}
	
	\renewcommand{\baselinestretch}{0.8}

\end{document}